\documentclass[10pt,twocolumn,letterpaper]{article}

\usepackage{cvpr}              

\usepackage{colortbl}
\usepackage{xcolor}
\definecolor{myGreen}{rgb}{0.3,0.9,0.1}
\definecolor{lightgreen}{rgb}{0.6,1,0.5}
\definecolor{darkorange}{rgb}{0.95,0.75,0.45} 
\definecolor{orange}{rgb}{1.0,0.85,0.5} 
\definecolor{lightorange}{rgb}{1.0,0.95,0.7}
\definecolor{lightgrey}{rgb}{0.95,0.95,0.95}
\definecolor{lightred}{rgb}{1.0,0.8,0.8}
\definecolor{lightblue}{rgb}{0.8,0.9,1.0}
\usepackage{multirow}
\usepackage{booktabs}
\usepackage{svg}
\usepackage{subcaption}
\definecolor{tabblue}{HTML}{1F77B4}
\definecolor{taborange}{HTML}{FF7F0E}
\definecolor{tabgreen}{HTML}{2CA02C}
\definecolor{tabred}{HTML}{D62728}
\definecolor{tabpurple}{HTML}{9467BD}
\definecolor{tabbrown}{HTML}{8C564B}
\definecolor{tabpink}{HTML}{E377C2}
\definecolor{tabgray}{HTML}{7F7F7F}
\definecolor{tabolive}{HTML}{BCBD22}
\definecolor{tabcyan}{HTML}{17BECF}

\definecolor{cvprblue}{rgb}{0.21,0.49,0.74}
\usepackage[pagebackref,breaklinks,colorlinks,allcolors=cvprblue]{hyperref}

\def\paperID{8} 
\def\confName{CVPR}
\def\confYear{2025}

\title{Machine Unlearning in Hyperbolic vs. Euclidean Multimodal Contrastive Learning: Adapting Alignment Calibration to MERU}

\author{
Àlex Pujol Vidal$^{1,3}$\\ \tt\small alexpv@create.aau.dk \and Kamal Nasrollahi$^{1,3,4}$ \\ \tt\small kna@milestone.dk \and Thomas B. Moeslund$^{1,4}$ \\ \tt\small tbm@create.aau.dk \and Sergio Escalera$^{1,2,4}$ \\ \tt\small sescalera@ub.edu\\
$^{1}$Aalborg University, Aalborg, Denmark\\
$^{2}$University of Barcelona and Computer Vision Center, Barcelona, Spain\\
$^{3}$Milestone Systems, Broendby, Denmark\\
$^{4}$Pioneer Center for Artificial Intelligence, Copenhagen, Denmark
}

\begin{document}
\maketitle
\begin{abstract}
Machine unlearning methods have become increasingly important for selective removal of harmful concepts in large multimodal models. While recent work has explored unlearning in Euclidean contrastive vision-language models, the effectiveness of concept removal in hyperbolic spaces remains unexplored. This paper explores this gap by adapting Alignment Calibration to MERU, a model that embeds images and text in hyperbolic space to better capture semantic hierarchies. Through systematic experiments and ablation studies, we demonstrate that hyperbolic geometry offers distinct advantages for concept removal, achieving lower accuracy at zero classification in forget set with reasonable performance on retain set, particularly when scaling to multiple concept removal. Our approach introduces hyperbolic-specific components including entailment calibration and norm regularization that leverage the unique properties of hyperbolic space. Comparative analysis with Euclidean models reveals fundamental differences in unlearning dynamics, with hyperbolic unlearning reorganizing the semantic hierarchy while Euclidean approaches merely disconnect cross-modal associations. These findings provide insights into the geometric properties that influence concept removal, setting a direction towards better control and safer deployment for large multimodal models. Code available at \url{https://github.com/alex-pv01/HAC}.
\end{abstract}    
\section{Introduction}
\label{sec:intro}

It is commonly said that 'we learn through contrast'. By comparing seemingly opposite concepts, we gain deeper understanding—a fundamental principle underlying contrastive learning algorithms. This approach, has emerged as a powerful framework for learning representations that effectively distinguish similar from dissimilar in machine learning systems \cite{contrastive-loss}.

The success of contrastive learning has led to the development of powerful multimodal models like CLIP, which aligns visual and textual representations in a shared Euclidean embedding space \cite{clip}. These models serve as foundational components for numerous downstream applications including zero-shot classification \cite{clip-zero-shot}, image generation \cite{clip-image}, and robotics \cite{clip-robot}. More recently, MERU \cite{meru} has extended this paradigm by embedding images and text in hyperbolic space, better capturing the hierarchical relationships inherent in visual-semantic data.

Despite their impressive capabilities, these models rely on massive datasets scraped from the internet, which often contain problematic content including private information, copyrighted material, and harmful data \cite{unsafe-data}. This raises significant ethical and legal concerns, as these models become widely deployed. Machine unlearning—the ability to selectively remove information from trained models—initially emerged in response to regulatory frameworks like GDPR's "right to be forgotten"\cite{gdpr}, but has since expanded to address a broader spectrum of challenges including privacy protection, safety enhancement, security concerns, fairness improvement, model robustness and copyright matters \cite{mu-survey, mu-doesnt-do}. 

In particular, concept removal (of NSFW or copyrighted content) has been increasingly relevant to ensure a safe deployment \cite{safe-clip}. While recent work has explored concept removal in Euclidean contrastive learning models like CLIP \cite{safe-clip, cliperase, ac, zeroshotunlearnlip, zeroshotunlearn-spaceadapt}, the unique geometry of hyperbolic spaces presents new challenges and opportunities that remain unexplored. The geometric distinction raises an important question: How does the underlying geometry affect the efficacy and dynamics of machine unlearning in contrastive learning models?


In this paper, we investigate the adaptation of Alignment Calibration (AC) to hyperbolic representations in MERU. We present a systematic comparison of unlearning effectiveness between CLIP and MERU models trained on identical data. Our contributions are threefold:
\begin{enumerate}
    \item Proposing Hyperbolic Alignment Calibration (HAC), an extension of AC to operate in hyperbolic space by reformulating its objective functions using hyperbolic distance and incorporating entailment constraints.
    \item Based on zero-shot classification evaluation, we conduct a comparative analysis of machine unlearning in Euclidean and hyperbolic contrastive learning spaces using AC and HAC, revealing that the hyperbolic method excels at removing the targeted concept at the cost of a decrease in performance on the retain set.
    \item Through visualization techniques, we complement our insights on how the geometries affect unlearning dynamics. Indicating that HAC may profit from the exponential expansion of the hyperbolic space to achieve better unlearning.
\end{enumerate}

This work bridges two important research areas: geometric representation learning and machine unlearning. Our study advances the understanding on how representation spaces encode and forget concepts, contributing to the theoretical foundations of machine unlearning for advanced vision language models. 

\section{Related work}
\label{sec:relwork}

\paragraph{Hyperbolic Representation Learning}
The choice of geometry for representation spaces has significant implications for the expressive power of learned embeddings. While most machine learning models operate in Euclidean space, recent work has explored alternative geometries, particularly hyperbolic geometry for data with hierarchical structure \cite{hyp-sur, hyp-sur-cv}.

Characterized by its negative curvature, a hyperbolic space provides a continuous analog of tree-like structures \cite{poincare-emb}. In hyperbolic space, the volume of a ball grows exponentially with its radius, allowing to embed hierarchical data with low distortion \cite{hyp-trade}. This property is valuable for representing data with natural hierarchies, such as taxonomies, social networks, and visual-semantic relationships \cite{poincare-emb, cones, meru}. These geometric insights have direct implications for multimodal contrastive learning, on how concepts are organized and related across modalities.
\vspace{-0.35cm}
\paragraph{Multimodal Contrastive Learning}
Contrastive learning emerged as a powerful approach for unsupervised representation learning, training models to distinguish between similar (positive) and dissimilar (negative) pairs of examples \cite{contrastive-loss}. 
This approach is formalized through the InfoNCE loss function \cite{infonce}. Given a batch of $N$ samples, where for every sample $z_i$, there is a positive pair $z_{k_i}\neq z_i$, then:
\begin{equation}\label{eq:infonce}
L_{\text{InfoNCE}} = \frac{-1}{N}\sum_{i=1}^N\log \frac{\exp(\text{sim}(z_i, z_{k_i})/\tau)}{\sum_{k=1, \; k \neq i}^N \exp(\text{sim}(z_i, z_k)/\tau)}
\end{equation}
where $\text{sim}(u, v)$ denotes cosine similarity, and $\tau$ is a temperature parameter.
CLIP \cite{clip} extended this approach to the multimodal domain by learning a dual-encoder architecture with separate visual and textual encoders that project images and their corresponding text descriptions into a shared Euclidean space. 
More recently, MERU \cite{meru} enhanced CLIP's representational power by embedding images and text in hyperbolic and using the negative hyperbolic distance as a similarity metric. It achieved improved performance on tasks involving hierarchical relationships while maintaining CLIP's zero-shot capabilities.

The transition from Euclidean to hyperbolic embeddings in multimodal contrastive learning raises important questions about how these different geometries affect the organization of concepts within the representation space.
\vspace{-0.35cm}
\paragraph{Machine Unlearning}
Machine Unlearning (MU), introduced by Cao and Yang \cite{mu-origins}, selectively removes the influence of specific training data from a trained model. While early research focused primarily on classification tasks \cite{mu-survey}, the rapid growth of generative AI has expanded the scope of MU to higher-order or latent information \cite{mu-doesnt-do, concept_prune, ablating_concepts}.


Particularly, several works on concept removal for contrastive vision-language models have recently emerged.
Poppi et al. \cite{safe-clip} addressed safety concerns by developing a method to diminish CLIP's sensitivity to NSFW content. Their approach is based on redirecting synthetic "unsafe" data to match "safe" data.
Kravets and Namboodiri \cite{zeroshotunlearnlip} applied Lipschitz regularization to achieve zero-shot class unlearning in CLIP. 
In subsequent work, Kravets and Namboodiri \cite{zeroshotunlearn-spaceadapt} demonstrated that class forgetting in CLIP can be accomplished without any visual data by adapting the shared vision-text space, making the forgetting process more efficient.
Yang et al. \cite{cliperase} introduced CLIPErase through three modules: a Forgetting Module that disrupts associations in the forget set, a Retention Module that preserves performance on the retain set, and a Consistency Module that maintains alignment with the original model.
Similarly, Wang et al. \cite{ac} proposed Alignment Calibration, which explicitly considers the properties of contrastive learning and introduce toward novel auditing metrics to verify unlearning.

While these approaches have shown promise for Euclidean-based models like CLIP, they do not account for the unique properties of hyperbolic geometry that are leveraged in models like MERU. 

Despite mentioned, no prior work has investigated how hyperbolic geometry affects concept removal in contrastive models. This gap is significant given hyperbolic space's natural capacity for representing hierarchical relationships, which could fundamentally alter how concepts are forgotten in multimodal representations.


\section{Method}
\subsection{Machine Unlearning for Concept Removal}\label{mu-concept}
Given a pre-trained contrastive model with image encoder $f_{\text{img}}$, text encoder $f_{\text{txt}}$, and a training dataset $\mathcal{D}$ with image-text pairs. Let $c$ be the concept to be removed, we define $\mathcal{D}_f$ as the subset containing instances that refer to $c$ and $\mathcal{D}_r = \mathcal{D} \setminus \mathcal{D}_f$ as the retain set. That is, if $c$ is the concept of "dog", $D_f$ consists of positive pairs of images of dogs and captions describing the image. The unlearning task is to find modify the original model and obtain $f^*_{\text{img}}, f^*_{\text{txt}}$ such that:
\begin{enumerate}
    \item The model no longer associates visual and textual representations of concept $c$. Formally, for a positive pair $(x_f, t_f) \in \mathcal{D}_f$ exists a pair $(x, t) \in \mathcal{D}_r$, \begin{equation}
        \text{sim}(f^*_{\text{img}}(x_f), f^*_{\text{txt}}(t_f)) \ll \text{sim}(f^*_{\text{img}}(x_f), f^*_{\text{txt}}(t)),
    \end{equation} and 
    \begin{equation}
        \text{sim}(f^*_{\text{img}}(x_f), f^*_{\text{txt}}(t_f)) \ll \text{sim}(f^*_{\text{img}}(x), f^*_{\text{txt}}(t_f)).
    \end{equation}
    \item Performance on other concepts in remains unaffected. 
    For positive pairs $(x, t) \in \mathcal{D}_r$, 
    \begin{equation}
        \text{sim}(f^*_{\text{img}}(x), f^*_{\text{txt}}(t)) \approx \text{sim}(f_{\text{img}}(x), f_{\text{txt}}(t)).
    \end{equation}
\end{enumerate}
Note that concept removal is a recent problem highlighted in literature and has no established formal definition \cite{mu-doesnt-do}. However, the above definition aligns with contrastive learning approaches on related work \cite{ac, zeroshotunlearnlip, zeroshotunlearn-spaceadapt}, where unlearning is measured by accuracy on zero-shot classification, as later discussed in \Cref{eval}.


\subsection{Alignment Calibration in Euclidean Space}

AC consists of a retaining loss and an unlearning loss that work together to selectively push apart image and text embeddings to forget, while preserving model performance \cite{ac}. Given a batch $\{(x^r_i,t^r_i)\}_{i=1}^N\subset\mathcal{D}_r$ and $\{(x^f_i,t^f_i)\}_{i=1}^N\subset \mathcal{D}_f$ from a dataset $\mathcal{D}$. Let $x'=f_\text{img}(x)$ and $t'=f_\text{txt}(t)$ denote image and text embeddings respectively. The retention loss consists of the usual loss for CLIP (see \cref{clip-sup} in supp. material), preserving model performance on $\mathcal{D}_r$:
\begin{align}\label{ac:ret}
    L_{\text{retain}} =& -\frac{1}{2N}\sum_{i=1}^{N} \Bigg[ \log \frac{\exp(\text{sim}(x^{'r}_i, t^{'r}_i)/\tau)}{\sum_{j=1}^{2N} \exp(\text{sim}(x^{'r}_i, t'_j)/\tau)} \\ 
    & + \log \frac{\exp(\text{sim}(x^{'r}_i, t^{'r}_i)/\tau)}{\sum_{j=1}^{2N} \exp(\text{sim}(x'_j, t^{'r}_i)/\tau)}\Bigg].
\end{align}

The unlearning loss consists of three components designed to disrupt associations in the forget set $\mathcal{D}_f$:
\begin{enumerate}
    \item Negative alignment calibration (\cref{ac:neg}), maximizes similarity between negative pairs in the forget set, confusing the model's understanding of $c$.
\begin{align}\label{ac:neg}
    L_{\text{neg}} = -\frac{1}{2N^2}\sum_{i=1}^{N}\sum_{j=1, j \neq i}^{N} \frac{\text{sim}(x^{'f}_i, t^{'f}_j) + \text{sim}(x^{'f}_j, t^{'f}_i)}{\tau}.
\end{align}
    \item Positive alignment calibration (\cref{ac:pos}), minimizes similarity between matching pairs in the forget set, directly attacking the concept association.
\begin{align}\label{ac:pos}
    L_{\text{pos}} = \frac{1}{N}\sum_{i=1}^{N}\text{sim}(x^{'f}_i, t^{'f}_i)/\tau.
\end{align}
    \item Performance preserving (\cref{ac:perf}), ensures that general model capabilities remain intact. Note that this corresponds to the denominator from the retention loss, \cref{ac:ret}, but also considering $D_f$.
\begin{align}\label{ac:perf}
    L_{\text{perf}} =& \frac{1}{2N}\sum_{i=1}^{N} \Bigg[\log\Big(\frac{1}{2N}\sum_{j=1}^{2N} \exp(\text{sim}(x^{'f}_i, t'_j)/\tau)\Big) \\
    &+ \log\Big(\frac{1}{2N}\sum_{j=1}^{2N} \exp(\text{sim}(x'_j, t^{'f}_i)/\tau)\Big)\Bigg]
\end{align}
\end{enumerate}

The combined unlearning loss is:
\begin{align}\label{ac:unlearn}
    L_{\text{forget}} = \alpha \cdot L_{\text{neg}} + \beta \cdot L_{\text{pos}} + \gamma \cdot L_{\text{perf}}.
\end{align}
And the final AC loss balances retention and unlearning objectives:\vspace{-0.3cm}
\begin{align}\label{ac:total}
    L_{\text{AC}} = L_{\text{retain}} + \varepsilon \cdot L_{\text{forget}},
\end{align}
where $\alpha$, $\beta$, $\gamma$, and $\varepsilon$ are hyperparameters controlling the relative importance of each component.

\subsection{Alignment Calibration to Hyperbolic Space}
HAC extends the AC approach to hyperbolic geometry by adapting each component of the loss function to operate in the Lorentz model of hyperbolic space. As in MERU's formulation (see \cref{meru-pre} in supp. material), we replace the cosine similarity with negative hyperbolic distance from component in AC. HAC incorporates entailment terms, introduced by Desai et al. \cite{meru}. We propose adapting the entailment to preserve and disrupt hierarchical relationships:\vspace{-0.1cm}
\begin{align}\label{hac:ret-ent}
    L_{\text{r-ent}} = \frac{1}{N}\sum_{i=1}^{N} \max(0, \text{ext}(x^{'r}_i, t^{'r}_i) - \text{aper}(t^{'r}_i)),
\end{align}\vspace{-0.3cm}
\begin{align}\label{hac:for-ent}
    L_{\text{f-ent}} = \frac{1}{N}\sum_{i=1}^{N} \max(0, \text{aper}(t^{'f}_i) - \text{ext}(x^{'f}_i, t^{'f}_i)),
\end{align}
where $\text{ext}(x, t)$ is the exterior angle between the text embedding $t$, and the image embedding $x$, and $\text{aper}(t)$ is the half-aperture of the entailment cone for $t$. The retain-entailment loss (\cref{hac:ret-ent}) preserves the hierarchical structure for concepts in the retain set, minimizing the angle difference when the embedded image $x^{'r}_i$ lies outside the entailment cone for the text $t^{'r}_i$, following the same logic as MERU. On the other had, the forget-entailment loss (\cref{hac:for-ent}) disrupts entailment relationships for texts and images of the concepts to be forgotten, minimizing the angle difference when $x^{'f}_i$ is inside the entailment cone of $t^{'f}_i$, forcing the images to move close to the edge of the cone. The combined HAC loss is:\vspace{-0.1cm}
\begin{align}\label{hac:total}
    L_{\text{HAC}} = L_{\text{retain}} + \varepsilon \cdot L_{\text{forget}} + \omega_r \cdot  L_{\text{r-ent}} + \omega_f \cdot L_{\text{f-ent}}.
\end{align}


Additionally, inspired by Li et al. \cite{ood}, who show that samples positioned close to the origin in hyperbolic space are likely to be out-of-distribution, we further propose adding a regularization term for the forgetting set that minimizes the hyperbolic norm of the image-text embeddings to be forgotten:
\vspace{-0.1cm}
\begin{align}\label{hac:reg}
    L_{\text{norm-reg}} = \frac{1}{N}\sum_{i=1}^{N} (||x^{'f}_i||_{\mathcal{L}} + ||t^{'f}_i||_{\mathcal{L}}),
\end{align}
where $||\cdot||_{\mathcal{L}}$ is the Lorentzian norm, see \cref{meru-pre}. This regularization pulls instances to remain closer to the origin, which in turn provides better numerical stability \cite{hyperclass}. The final HAC loss with regularization is:
\vspace{-0.1cm}
\begin{align}\label{hac:total-reg}
    L_{\text{HAC-reg}} = L_{\text{HAC}} + \lambda \cdot  L_{\text{norm-reg}}
\end{align}


\subsection{Implementation Details}
We implement both AC and HAC, building on the official CLIP and MERU codebases. For optimization, we use Adam \cite{adam} with learning rate $\eta = 5\cdot10^{-5}$, weight decay $\lambda = 10^{-5}$, and cosine learning rate decay over 15,000 iterations. We train with a batch size of 320 across all experiments. To ensure numerical stability in hyperbolic space computations, we apply gradient clipping with a maximum norm of 1.0. For unlearning, we train on a subset of RedCaps dataset \cite{redcaps}, running experiments on a NVIDIA A100 GPU. We evaluate both models quantitatively and qualitatively on publicly available datasets \ref{eval}.
\section{Experiments}
\subsection{Experimental Setup}

To investigate the impact of removing concepts from contrastive representation spaces, we focus on selectively disrupting the semantic relationship between image and text of concepts, such as dogs, cats, food, and plants. Both CLIP and MERU models have ViT-S \cite{vits} as image encoder and a 12-layer, 512 wide Trasnformer as text encoder \cite{clip}. They are pretrained on the RedCaps dataset and used as the orginal models\footnote{Models availabe in MERU's \href{https://github.com/facebookresearch/meru?tab=readme-ov-file}{GitHub repository}} that require unlearning.

During experiments only a subset of 7M image-text pairs from the original RedCaps dataset was accessible. This, referred to as RedCaps2, serves as the original dataset. Instances related to dogs, cats, food, and plants are categorized into four distinct high-level semantic concepts, defining the different forget sets, see \cref{forget-set} in supp. material for more details on the set up forget sets.
The unlearning procedure operates on balanced mini-batches with $N$ image-text pairs from the retain set and $N$ pairs from the unlearn set, resulting in a batch size of $2N$, with $N=160$. 


\subsection{Evaluation}\label{eval}
\paragraph{Zero-shot image classification}
The evaluation constructs prompts of the form "a picture of a [CLASS]" for each target class. Each test image embedding is compared against all class prompt embeddings using the appropriate similarity metric. The class with maximum similarity is selected as the prediction. Both accuracy on retained classes (R-acc) and forgotten classes (F-acc) are reported. A high F-acc indicates that the text-image embeddings are still close enough capturing the undesired semantic relationship in the latent space. 
\vspace{-0.35cm}
\paragraph{Latent Space Visualizations}
Image and text embeddings are visualized using both T-SNE \cite{tsne} and hyperbolic T-SNE \cite{cosne}. The latter is an extension of T-SNE which uses hyperbolic student's t-distribution to capture the hierarchical relationships between embeddings. These visualizations may reveal structural changes in the representation space after unlearning. We look particularly if there is a distortion in the alignment between the image and text embedding of the concepts to forget. 
\vspace{-0.35cm}
\paragraph{Datasets}\label{datasets}
For evaluation, we selected a subset of the 6 datasets used in the original MERU paper \citep{meru}, focusing on those containing classes related to our target concepts for unlearning. 
Additionally, for latent space visualization purposes we use a subset of COCO \cite{coco}, containing images of a selected group of classes.

\subsection{Ablation Studies}\label{sec:ablation}
\paragraph{Alignment Calibration Components}\label{sec:ablation_ac}
We assess the impact of the different alignment components on unlearning performance. We vary the weights from Eq. \ref{ac:unlearn} ($\alpha, \beta, \gamma\in [0, 1]$), unlearning a single concept (dogs) in both CLIP and MERU. As baselines, we consider the orignal CLIP and MERU (o-C and o-M), both models finetuned only with the retain dataset (f-C-R and f-M-R), and both models finetuned on the entire dataset (f-C and f-M).
\vspace{-0.35cm}
\paragraph{Hyperbolic-Specific Parameters}\label{sec:ablation_hyperbolic}
We ablate parameters unique to HAC: entailment weights ($\omega_r, \omega_u \in [0, 1]$) and hyperbolic norm regularization ($\lambda\in[0,2]$). We analyze how these hyperbolic-specific factors influence unlearning, evaluating on zero-shot classification for CIFAR-10.
\vspace{-0.35cm}
\paragraph{Scaling Analysis}\label{sec:ablation_scale}
We evaluate on diverse datasets and three experiments. Experiment A, consists of removing the concept-class dogs alone. Removing dogs serves as an ideal initial test case, since it has a strong semantic relationship with other concepts to retain (e.g., horses, cats). Experiment B extends this by removing both dogs and cats. Testing whether the unlearning mechanism can handle multiple related concepts simultaneously. Finally, Experiment C consists of unlearning dogs, cats, food, and plants, to test how each geometry handles large-scale concept removal without catastrophic forgetting of non-targeted concepts. 

\section{Results and Discussion}

\subsection{Alignment Calibration Components}

We begin our analysis by examining how changing the key components of AC affects unlearning performance. Table \ref{tab:beta_ablation} presents the effects of varying positive alignment calibration ($\beta$). For better numerical stability of the experiments, we introduce a small regularization norm with $\lambda=0.1$ for HAC. We observe that increasing $\beta$ leads to a significant reduction in forget accuracy (F-acc) in both AC and HAC, confirming that this component effectively pushes positive image-text pairs further apart in their respective embedding spaces and leading to zero-shot misclassifications. This is consistent with previous research showing that controlled modification of alignment between positive pairs can lead to selective forgetting \citep{cliperase, ac}. However, we observe a dramatic difference in unlearning effectiveness. In HAC, introducing even a small $\beta$, it causes forget accuracy to plummet on CIFAR-10. This suggests that hyperbolic geometry amplifies the effect of positive alignment calibration, potentially due to its natural exponential expansion.
The cost of this enhanced forgetting capability is reflected in retain accuracy (R-acc). HAC experiences a more substantial drop in retain performance compared to AC when $\beta$ is increased. Despite this trade-off, the relative difference between the drop in retain accuracy versus forget accuracy is more favorable in hyperbolic space, indicating superior unlearning efficiency. 
\begin{table}[t]
\centering
\caption{Zero-shot classification accuracy in retain and forget sets, varying positive alignment calibration. Largest difference in retain and forget performance in \textcolor{myGreen}{\textbf{green}}. Best value for each column and geometry in \textbf{bold}.}
\label{tab:beta_ablation}
\renewcommand{\arraystretch}{1.0}
\setlength{\tabcolsep}{3.5pt}
\begin{tabular}{c@{\hspace{1pt}}cc@{\hspace{4pt}}cc@{\hspace{4pt}}cc}
\specialrule{1.5pt}{0pt}{0pt}
\multirow{2}{*}{Method} & \multicolumn{2}{c}{Weights} & \multicolumn{2}{c}{CIFAR-10} & \multicolumn{2}{c}{O-IIIT Pets} \\
\cmidrule(lr){2-3} \cmidrule(lr){4-5} \cmidrule(lr){6-7}
& $\alpha,\gamma$ & $\beta$ & R-acc$\uparrow$ & F-acc$\downarrow$ & R-acc$\uparrow$ & F-acc$\downarrow$ \\
\specialrule{1.5pt}{0pt}{0pt}
\multirow{4}{*}{AC} & \multirow{4}{*}{0.75} & 0 & \textbf{60.5} & 45.6 & 73.6 & 66.2 \\
&  & 0.25 & 60.3 & 31.7 & 73.5 & 48.7 \\
&  & 0.5 & \cellcolor{lightgreen}58.7 & \cellcolor{lightgreen}\textbf{21.2} & \textbf{74.9} & 31.5 \\
&  & 0.75 & 58.4 & 24.9 & \cellcolor{lightgreen}73.9 & \cellcolor{lightgreen}\textbf{24.6} \\
\midrule
f-C & \multirow{3}{*}{0} & \multirow{3}{*}{0} & 58.9 & 48.1 & 74.6 & 69.9 \\
f-C-R &  &  & 60.6 & 63.6 & 72.3 & 72.2 \\
O-C &  &  & 59.4 & 66.4 & 74.6 & 73.2 \\
\specialrule{1.5pt}{0pt}{0pt}
\multirow{4}{*}{HAC} & \multirow{4}{*}{0.5} & 0 & 55.2 & 73.6 & 74.8 & 63.9 \\
&  & 0.25 & 34.7 & \textbf{0.0} & 62.1 & \textbf{15.8} \\
&  & 0.5 & \cellcolor{lightgreen}49.9 & \cellcolor{lightgreen}\textbf{0.0} & \cellcolor{lightgreen}69.6 & \cellcolor{lightgreen}17.9 \\
&  & 0.75 & 39.6 & 0.03 & 63.9 & 16.1 \\
\midrule
f-M & \multirow{3}{*}{0} & \multirow{3}{*}{0} & \textbf{56.4} & 73.0 & \textbf{75.2} & 65.9 \\
f-M-R &  &  & 41.7 & 95.7 & 71.8 & 65.8 \\
O-M &  &  & 38.1 & 94.6 & 72.0 & 70.8 \\
\specialrule{1.5pt}{0pt}{0pt}
\end{tabular}
\end{table}



On the other side, table \ref{tab:ag_ablation} examines the effects of varying negative alignment calibration ($\alpha$) and performance preserving ($\gamma$) weights while maintaining a fixed positive alignment calibration. For AC, we observe that the retain accuracy (R-acc) remains relatively stable across different $\alpha$ and $\gamma$ values, while forget accuracy (F-acc) shows more substantial fluctuations. This suggests that in Euclidean space, modifying $\alpha$ and $\gamma$ has a more pronounced effect on the model's ability forget the selected concept. In contrast, HAC demonstrates remarkably consistent forgetting capabilities across all parameter configurations, maintaining near-zero forget accuracy regardless of $\alpha$ and $\gamma$ values, while dampening retain accuracy. This remarkable behavior suggests that in hyperbolic space, once the positive alignment calibration establishes a forgetting directive, it dominates the unlearning process to such an extent that adjustments to other components extend the disruption to the retain set.


\begin{table}[t]
\centering
\caption{Zero-shot classification accuracy in retain forget sets, varying negative alignment calibration and performance preserving. Largest difference in retain and forget performance in \textcolor{myGreen}{\textbf{green}}. Best value for each column and geometry in \textbf{bold}.}
\label{tab:ag_ablation}
\renewcommand{\arraystretch}{1.0}
\setlength{\tabcolsep}{4pt}
\begin{tabular}{c@{\hspace{2pt}}cc@{\hspace{5pt}}cc@{\hspace{5pt}}cc}
\specialrule{1.5pt}{0pt}{0pt}
\multirow{2}{*}{Method} & \multicolumn{2}{c}{Weights} & \multicolumn{2}{c}{CIFAR-10} & \multicolumn{2}{c}{O-IIIT Pets} \\
\cmidrule(lr){2-3} \cmidrule(lr){4-5} \cmidrule(lr){6-7}
& $\alpha,\gamma$ & $\beta$ & R-acc$\uparrow$ & F-acc$\downarrow$ & R-acc$\uparrow$ & F-acc$\downarrow$ \\
\specialrule{1.5pt}{0pt}{0pt}
\multirow{3}{*}{AC} & 0.5 & \multirow{3}{*}{0.5} & \textbf{58.8} & 24.0 & 74.6 & 32.9 \\
& 0.75 &  & \cellcolor{lightgreen}58.7 & \cellcolor{lightgreen}\textbf{21.2} & \cellcolor{lightgreen}\textbf{74.9} & \cellcolor{lightgreen}\textbf{31.5} \\
& 1 &  & 57.2 & 27.4 & 73.6 & 41.5 \\
\midrule
\multirow{3}{*}{HAC} & 0.5 & \multirow{3}{*}{0.5} & \cellcolor{lightgreen}\textbf{49.9} & \cellcolor{lightgreen}\textbf{0.0} & \cellcolor{lightgreen}\textbf{69.6} & \cellcolor{lightgreen}\textbf{17.9} \\
& 0.75 &  & 40.7 & 0.02 & 67.5 & 18.0 \\
& 1 &  & 42.7 & 0.04 & 68.6 & 19.8 \\
\specialrule{1.5pt}{0pt}{0pt}
\end{tabular}
\end{table}


\subsection{Hyperbolic-Specific Parameters}

We now explore parameters that are unique to the hyperbolic geometry. Table \ref{tab:ent} illustrates the impact of varying entailment loss weights on HAC performance. When AC components are deactivated ($\varepsilon = 0$), we observe not only a failure to achieve unlearning but actually an increase in forget accuracy on CIFAR-10. This finding suggests that entailment losses alone may inadvertently strengthen the association between text and image embeddings of the target concept, increasing their relative similarity, instead of decreasing it.

\begin{table}[t]
\centering
\caption{Zero-shot classification accuracy in retain and forget sets, varying the weight entailment losses. Largest difference in retain and forget performance in \textcolor{myGreen}{\textbf{green}}. Best value for each column and geometry in \textbf{bold}.}
\label{tab:ent}
\renewcommand{\arraystretch}{1.0}
\setlength{\tabcolsep}{4pt}
\begin{tabular}{ccc@{\hspace{5pt}}cc@{\hspace{5pt}}cc}
\specialrule{1.5pt}{0pt}{0pt}
\multicolumn{3}{c}{Weights} & \multicolumn{2}{c}{CIFAR-10} & \multicolumn{2}{c}{O-IIIT Pets} \\
\cmidrule(lr){1-3} \cmidrule(lr){4-5} \cmidrule(lr){6-7}
$\epsilon$ & $\omega_r$ & $\omega_f$ & R-acc$\uparrow$ & F-acc$\downarrow$ & R-acc$\uparrow$ & F-acc$\downarrow$ \\
\specialrule{1.5pt}{0pt}{0pt}
\multirow{2}{*}{0} & 0.2 & 1.0 & \textbf{56.3} & 73 & \textbf{74.9} & 66.6 \\
 & 1.0 & 0.2 & 47.2 & 85.9 & 68.5 & 64.6 \\
\multirow{2}{*}{0.05} & 0.2 & 1.0 & 39.9 & \textbf{0.0} & \cellcolor{lightgreen}60.7 & \cellcolor{lightgreen}\textbf{0.08} \\
 & 1.0 & 0.2 & 49.6 & 37.0 & 40.1 & 0.10 \\
\multirow{2}{*}{0.1} & 0.2 & 1.0 & \cellcolor{lightgreen}52.7 & \cellcolor{lightgreen}\textbf{0.0} & 67.9 & 15.1 \\
 & 1.0 & 0.2 & 44.0 & 48.7 & 56.8 & 18.6 \\
\specialrule{1.5pt}{0pt}{0pt}
\end{tabular}
\end{table}


The introduction of AC components ($\varepsilon > 0$) dramatically changes this dynamic. Furthermore, when prioritizing the forget entailment ($\omega_f = 1.0$) complete forgetting (0\% F-acc) on CIFAR-10 is achieved. This demonstrates the synergistic effect between alignment calibration and entailment losses in hyperbolic space. Similar patterns emerge for O-IIIT Pets, yielding near-perfect forgetting while maintaining reasonable retain accuracy.

The balance between retain and forget entailment weights ($\omega_r$ and $\omega_f$) proves crucial for optimal unlearning. When $\omega_r > \omega_f$, we observe significantly higher forget accuracy across all configurations, indicating incomplete concept removal. This supports the theoretical understanding that in hyperbolic space, the precise positioning of embeddings within entailment cones strongly influences model behavior \citep{meru}.

Finally, beyond providing better stability, the regularization term actively improves unlearning quality, as shown in Table \ref{tab:norm}. With $\lambda=0.5$, we largest retain-forget trade-off across both datasets. This regularization effect is particularly beneficial for retaining accuracy illustrated with O-IIIT Pets, where at the cost of increasing forgetting accuracy, we observe a strong retain performance. The mechanism behind this improvement involves pulling text embeddings for the forgotten concept toward the origin of the hyperboloid, causing other images that were previously misclassified, to be reclassified under other categories, provably their category, increasing retain accuracy. This is especially important for fine-grained classification in O-IIIT Pets, where the regularization helps disentangle dogs from similar categories like cats.

\begin{table}[t]
\centering
\caption{Zero-shot classification accuracy in retain and forget sets, varying the hyperbolic norm regularization. Largest difference in retain and forget performance in \textcolor{myGreen}{\textbf{green}}. Best value for each column and geometry in \textbf{bold}.}
\label{tab:norm}
\renewcommand{\arraystretch}{1.0}
\setlength{\tabcolsep}{5pt}
\begin{tabular}{l@{\hspace{2pt}}c@{\hspace{5pt}}cc@{\hspace{5pt}}cc}
\specialrule{1.5pt}{0pt}{0pt}
\multirow{2}{*}{Method} & Weight & \multicolumn{2}{c}{CIFAR-10} & \multicolumn{2}{c}{O-IIIT Pets} \\
\cmidrule(lr){3-4} \cmidrule(lr){5-6}
& $\lambda$ & R-acc$\uparrow$ & F-acc$\downarrow$ & R-acc$\uparrow$ & F-acc$\downarrow$ \\
\specialrule{1.5pt}{0pt}{0pt}
HAC & 0 & 52.0 & 13.0 & 46.3 & \textbf{0.04} \\
\midrule
\multirow{3}{*}{HAC-reg} & 0.1 & 52.7 & \textbf{0.0} & 67.9 & 15.1 \\
& 0.5 & \cellcolor{lightgreen}54.0 & \cellcolor{lightgreen}\textbf{0.0} & \cellcolor{lightgreen}66.3 & \cellcolor{lightgreen}10.8 \\
& 2.0 & \textbf{56.8} & 49.2 & \textbf{74.8} & 35.9 \\
\specialrule{1.5pt}{0pt}{0pt}
\end{tabular}
\end{table}

However, excessive regularization ($\lambda=2.0$) diminishes unlearning effectiveness, as evidenced by the significant increase in forget accuracy. This indicates that while regularization is essential for hyperbolic unlearning, its strength must be carefully calibrated to achieve the optimal balance between concept removal and retention.


\subsection{Scaling Analysis}

Fixing hyperparametes to the combination found with largest forget-retain difference, for both AC and HAC, we now investigate how these approaches scale when progressively increasing the complexity of the unlearning task. Table \ref{tab:scaling_analysis} presents a comprehensive comparison across multiple datasets when unlearning: (A) only "dog"; (B) "dog" and "cat"; and (C) "dog", "cat", "food", and "plants". 

\begin{table*}[t]
\centering
\caption{Zero-shot classification accuracy in retain set (R-acc) and forget set (F-acc), across different tasks, after unlearning: (A) "dog"; (B) "dog" and "cat"; (C) "dog", "cat", "food" and "plant". We report results for both CLIP and MERU after alignment calibration using the optimal configuration from Section~\ref{sec:ablation_ac}. Values in \textbf{bold} indicate better at retaining or unlearning across A, B and C. A blank space - indicate that for that experiment and dataset there is no forget or retain set.}
\label{tab:scaling_analysis}
\resizebox{\textwidth}{!}{
\begin{tabular}{lcccccccccccccc}
\specialrule{1.5pt}{0pt}{0pt}
\multirow{2}{*}{Task} & \multirow{2}{*}{Method} & \multirow{2}{*}{\begin{tabular}[c]{@{}c@{}}Unlearn\\Set\end{tabular}} & \multicolumn{2}{c}{CIFAR-10\cite{cifar10}} & \multicolumn{2}{c}{CIFAR-100\cite{cifar100}} & \multicolumn{2}{c}{STL-10\cite{stl10}} & \multicolumn{2}{c}{O-IIIT Pets\cite{pets}} & \multicolumn{2}{c}{Food101\cite{food101}} & \multicolumn{2}{c}{Flowers102\cite{flowers102}} \\
\cmidrule(lr){4-5} \cmidrule(lr){6-7} \cmidrule(lr){8-9} \cmidrule(lr){10-11} \cmidrule(lr){12-13} \cmidrule(lr){14-15}
& & & R-acc$\uparrow$ & F-acc$\downarrow$ & R-acc$\uparrow$ & F-acc$\downarrow$ & R-acc$\uparrow$ & F-acc$\downarrow$ & R-acc$\uparrow$ & F-acc$\downarrow$ & R-acc$\uparrow$ & F-acc$\downarrow$ & R-acc$\uparrow$ & F-acc$\downarrow$ \\
\specialrule{1.5pt}{0pt}{0pt}
 & & A & \textbf{58.7} & 21.2 & \textbf{27.9} & - & \textbf{88.1} & 83.1 & \textbf{74.9} & 31.5 & \textbf{72.4} & - & \textbf{44.7} & - \\
 & & B & \textbf{90.3} & 71.4 & \textbf{26.6} & - & \textbf{90.3} & 71.4 & - & 53.4 & \textbf{72.5} & - & \textbf{45.0} & - \\
 & \multirow{-3}{*}{AC} & C & \textbf{90.0} & 77.0 & \textbf{23.4} & 57.2 & \textbf{90.0} & 77.0 & - & 64.0 & - & 0.16 & - & 19.2 \\
 \cmidrule(l){2-15}
 &  &  A & 54.0 & \textbf{0.0} & 20.6 & - & 84.3 & \textbf{38.0}& 66.3 & \textbf{10.8} & 67.6 & - & 40.1 & -  \\
 & & B & 83.5 & \textbf{2.1} & 21.8 & - & 83.5 & \textbf{2.1} & - & \textbf{25.7} & 59.6 & - & 36.4& - \\
 \multirow{-6}{*}{\begin{tabular}[c]{@{}l@{}}Zero-shot\\Classification\end{tabular}}& \multirow{-3}{*}{HAC-reg}& C & 82.7 & \textbf{22.1} & 18.8 & \textbf{21.6} & 82.7 & \textbf{22.1} & - & \textbf{28.7} & - & \textbf{0.08} & - & \textbf{0.04} \\
\specialrule{1.5pt}{0pt}{0pt}
\end{tabular}
}
\end{table*}

We observe a clear trade-off pattern between retain accuracy and forget accuracy across all experiments. AC consistently maintains higher retain accuracy across all datasets and concept sets compared to HAC. However, this advantage comes at a significant cost: AC's forget accuracy increases dramatically as the unlearning task scales, indicating incomplete concept removal and lower scaling capabilities. 
In stark contrast, HAC demonstrates superior forgetting capabilities across all scaling scenarios. Even when tasked with unlearning four diverse concepts simultaneously (experiment C), HAC maintains remarkably low forget accuracies. Suggesting that HAC's leverages the hierarchical hyperbolic structure to effectively remove multiple concepts from the embedding space without cross-interference.

These experiments reveal that HAC offers more robust concept removal at scale, while AC better preserves performance on retained concepts. This fundamental trade-off highlights the complementary strengths of each geometric approach and suggests that the choice between them should be guided by whether the priority is complete concept removal or minimal disruption to retained knowledge. \Cref{confusion-sup} in supplementary material complements our findings showcasing the confusion matrices foer CIFAR-10.

\section{Latent Space Visualizations}\label{sec:qual}

\begin{figure*}[htbp]
    \centering
    \begin{subfigure}[b]{0.23\textwidth}
        \centering
        \includegraphics[width=\textwidth]{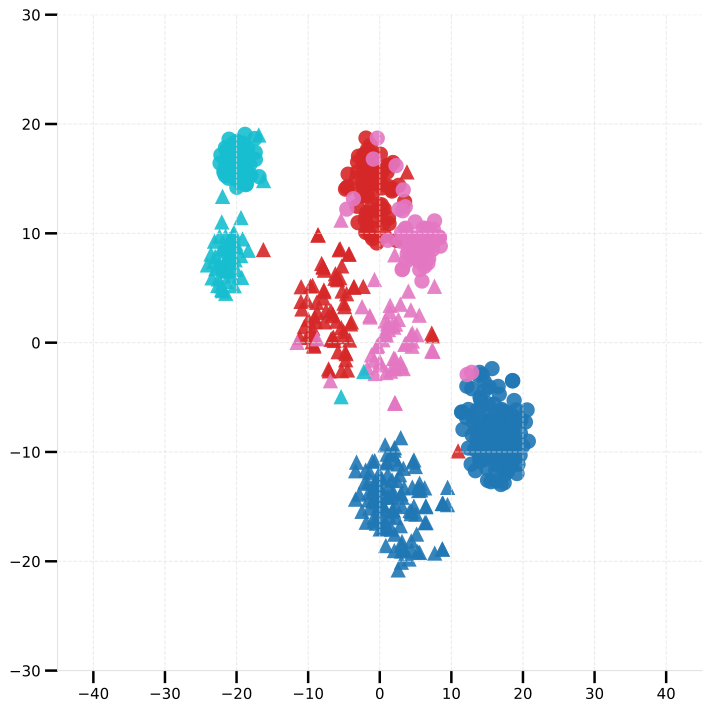}
        \caption{Original CLIP}
        \label{fig:sub1}
    \end{subfigure}
    \hfill
    \begin{subfigure}[b]{0.23\textwidth}
        \centering
        \includegraphics[width=\textwidth]{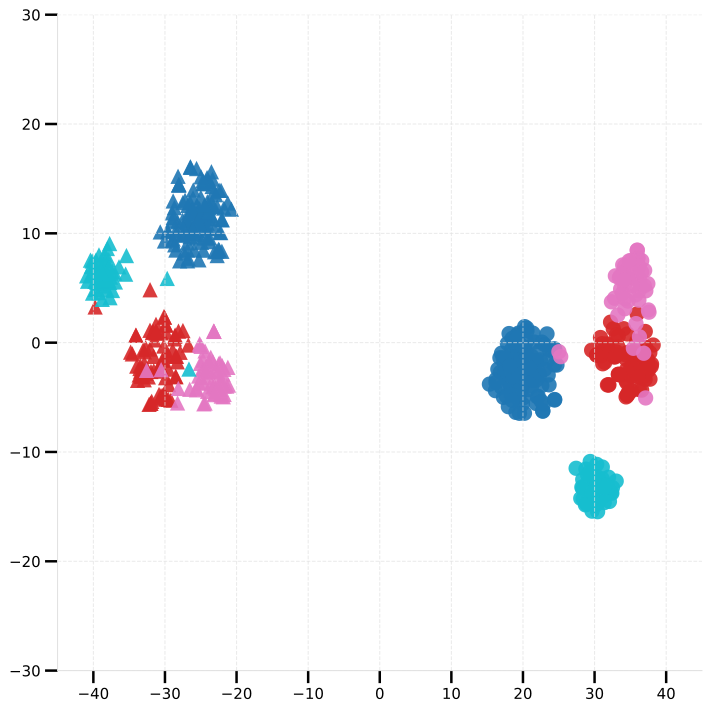}
        \caption{Unlearned CLIP}
        \label{fig:sub2}
    \end{subfigure}
    \hfill
    \begin{subfigure}[b]{0.23\textwidth}
        \centering
        \includegraphics[width=\textwidth]{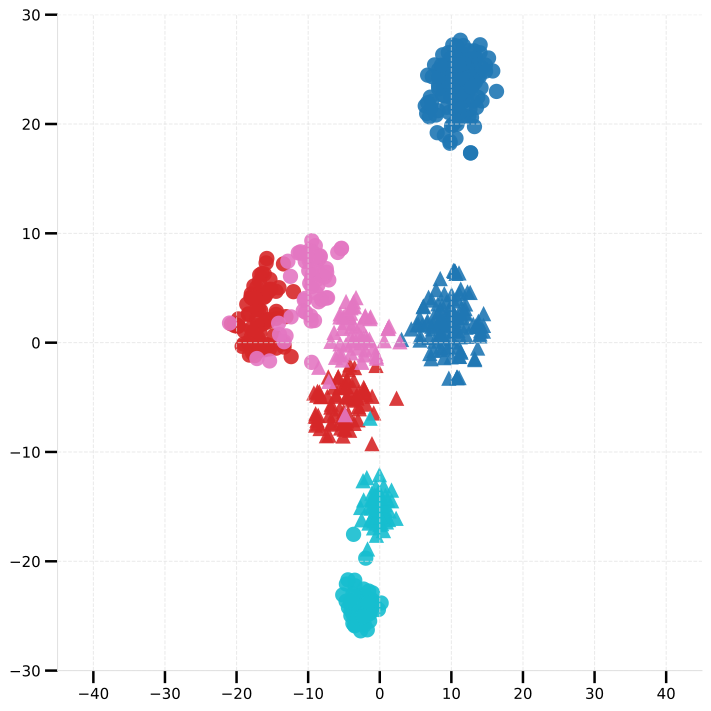}
        \caption{Original MERU}
        \label{fig:sub3}
    \end{subfigure}
    \hfill
    \begin{subfigure}[b]{0.23\textwidth}
        \centering
        \includegraphics[width=\textwidth]{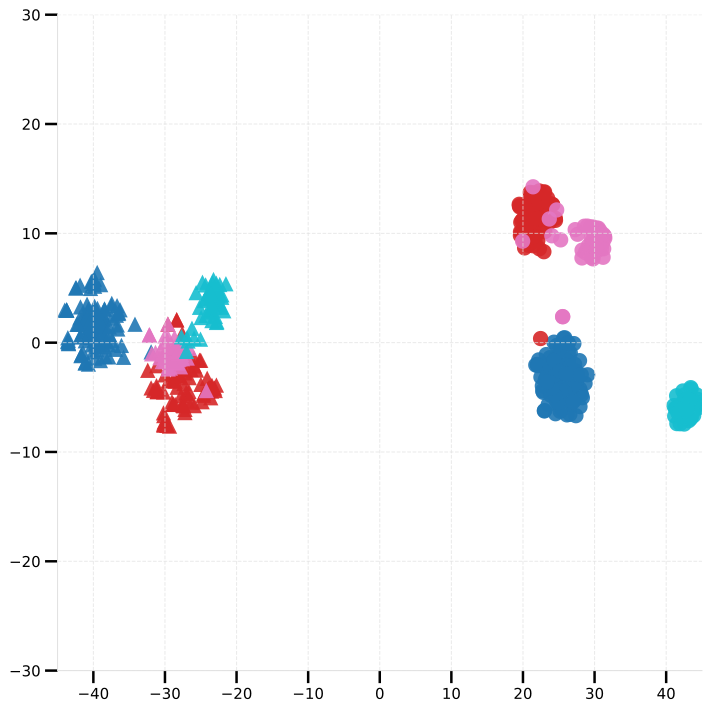}
        \caption{Unlearned MERU}
        \label{fig:sub4}
    \end{subfigure}
    \caption{Latent space visualizations with T-SNE of CLIP and MERU before and after removing the concept "dog". $\triangle$ refer to text embeddings, $\circ$ to image embeddings, and colors to \textcolor{tabpink}{\textbf{dogs}}, \textcolor{tabred}{\textbf{cats}}, \textcolor{tabcyan}{\textbf{pizzas}}, and \textcolor{tabblue}{\textbf{buses}}.}
    \label{fig:all}
\end{figure*}

\begin{figure}[htbp]
    \centering
    \begin{subfigure}[b]{0.22\textwidth}
        \centering
        \includegraphics[width=\textwidth]{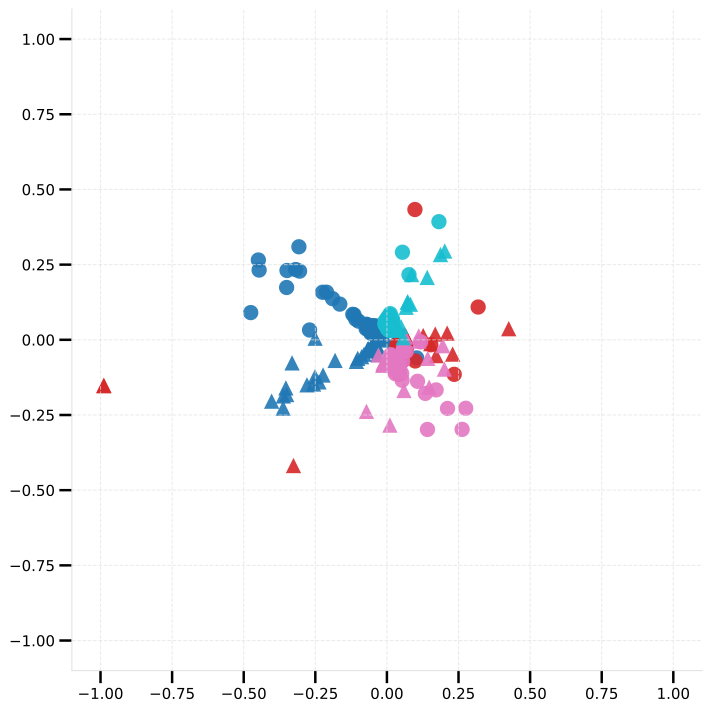}
        \caption{Original MERU}
        \label{fig:sub1}
    \end{subfigure}
    \hfill
    \begin{subfigure}[b]{0.22\textwidth}
        \centering
        \includegraphics[width=\textwidth]{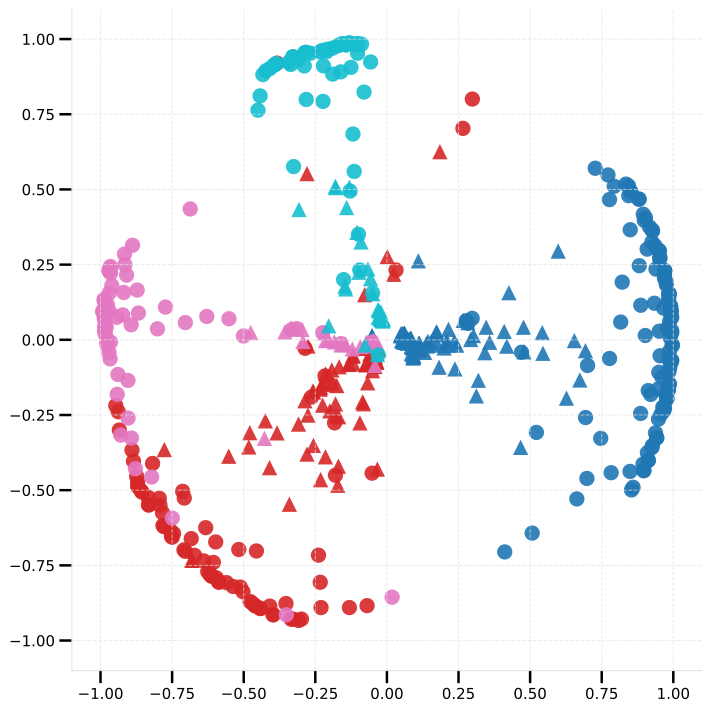}
        \caption{Unlearned MERU}
        \label{fig:sub2}
    \end{subfigure}
    \caption{Latent space visualizations with hyperbolic T-SNE of MERU before and after removing the concept "dog". $\triangle$ refer to text embeddings, $\circ$ to image embeddings, and colors to \textcolor{tabpink}{\textbf{dogs}}, \textcolor{tabred}{\textbf{cats}}, \textcolor{tabcyan}{\textbf{pizzas}}, and \textcolor{tabblue}{\textbf{buses}}.}
    \label{fig:htsne}
\end{figure}

To gain deeper insights into how geometric properties affect concept unlearning, we visualize the embedding spaces of both CLIP and MERU before and after removing the concept "dog". Figures \ref{fig:all} and \ref{fig:htsne} provide compelling visual evidence of the distinctive unlearning mechanisms in Euclidean versus hyperbolic spaces.


The T-SNE visualizations in Figure \ref{fig:all} reveal several key patterns. First, both AC and HAC induce a general separation between image and text representations after unlearning, indicating that enlarging the modality gap is a common consequence of unlearning regardless of geometry. However, the degree and nature of this separation differ between models. In the Euclidean space of CLIP (Figures \ref{fig:all}a-b), both text and image embeddings maintain their class-wise clustering structure after unlearning. In contrast, MERU's hyperbolic space (Figures \ref{fig:all}c-d) exhibits a more profound reorganization of semantic relationships. While image embeddings preserve their class separation, text embeddings show significant restructuring, with dog text representations (pink triangles) migrating toward the region occupied by cat text embeddings (red triangles). This visualization corroborates our quantitative findings that hyperbolic geometry facilitates complete concept removal by leveraging semantic hierarchies—in this case, potentially positioning "dog" as a subconcept of a more general "animal" category that includes cats. 
However, this poses a potential limitation of hyperbolic unlearning: while effective at repositioning target concepts, the structural interdependence of the hierarchical space means that semantically adjacent concepts (like cats when removing dogs) experience collateral disruption in their text embeddings, potentially explaining the lower retention accuracy observed in HAC compared to AC.

The hyperbolic T-SNE visualizations in Figure \ref{fig:htsne} provide additional insights into MERU's unlearning dynamics. Following unlearning, we observe a dramatic expansion of the embedding space, with image representations pushed substantially farther from the origin while text embeddings remain relatively close to it. This pattern aligns with hyperbolic geometry's exponential expansion property and illustrates how HAC exploits this characteristic. Particularly notable is the positioning of dog text embeddings near the origin after unlearning, effectively placing them "behind" cat text embeddings from the perspective of dog images. This spatial reorganization explains the superior unlearning results evaluated in zero-shot classification, dog images now find stronger alignment with text embeddings of other classes because their original text concept has been repositioned closer to the origin within the semantic hierarchy. Complementary illustrations can be found in \Cref{extra-sup} in supplementary material.

These visualizations demonstrate that while both geometric approaches achieve concept unlearning, they do so through fundamentally different mechanisms: Euclidean unlearning primarily disconnects cross-modal associations while preserving class structure, whereas hyperbolic unlearning reorganizes the semantic hierarchy itself, repositioning forgotten concepts within the taxonomic structure of the embedding space.

\section{Limitations}
While our work provides valuable insights into hyperbolic unlearning for contrastive models, several key limitations must be acknowledged. First, how to properly define concept boundaries for removal and defining a forgetting dataset remains an open problem in MU \citep{mu-doesnt-do}. To mitigate this issue, in this paper, we focus on concepts that correspond to discrete classes in common image-text datasets, see \cref{forget-set} for more details. However, given $D_f$, the formulation can apply to more abstract concepts as well.

Our interpretation of concept removal, disrupting text-image alignment and measuring effects through downstream tasks, aligns with related work \citep{ac, cliperase, zeroshotunlearnlip, zeroshotunlearn-spaceadapt}. However, zero-shot classification relies on simple sentences that may not capture the full semantic complexity of visual concepts, potentially limiting our ability to precisely measure text-image alignment degradation.


The inherent numerical instability of hyperbolic space presents additional challenges. The experiments required gradient clipping beyond MERU's original training pipeline to prevent exploding gradients caused by positive alignment calibration interacting with hyperbolic expansion. A more thorough analysis of numerical stability specifically for unlearning tasks is needed, as these instabilities might actually be leveraged to more effectively disrupt text-image alignments for target concepts.
For more robust conclusions, experiments should be conducted with multiple random seeds. The experiments used a single seed (0) across all conditions. While this allows us to establish initial insights into concept removal for hyperbolic contrastive learning in vision-language models, additional experimentation with varied seeds would strengthen the reliability of our findings.

Finally, the latent space visualizations rely on T-SNE and hyperbolic T-SNE, which, while informative, introduce distortions when projecting high-dimensional embeddings to 2D space, potentially ignoring important geometric properties of the original hyperbolic manifold and making it difficult to fully capture the complex hierarchical relationships that emerge during the unlearning process.

Despite these limitations, our work represents an important first step in understanding the geometric implications of concept unlearning in hyperbolic contrastive vision-language models, providing a foundation for future research in this emerging area.

\section{Conclusion}
In this work, we adapted Alignment Calibration for concept removal in hyperbolic contrastive learning. The comparison between Euclidean and hyperbolic geometries reveals that while AC better preserves retained concepts with high retain accuracy, HAC achieves more concept removal with low forget accuracy, particularly when scaling to multiple related concepts. This difference stems from hyperbolic geometry's hierarchical structure, which enables precise manipulation of semantic relationships through entailment weights and norm regularization. Visualizations provide insights that hyperbolic unlearning reorganizes the semantic hierarchy itself, rather than merely separating cross-modal instances. These findings underscore the importance of geometry in multimodal representation learning and provide a foundation for further research into hyperbolic unlearning for vision-language models. Future work should focus on developing specialized metrics and interpretability techniques for hyperbolic concept removal, addressing numerical stability challenges, and exploring how these approaches transfer to larger multimodal models.
\section{Acknowledgements}
This work has been supported by Milestone Research Program at AAU; the Responsible AI for Value Creation project (REPAI); the Spanish project PID2022-136436NB-I00; and ICREA under the ICREA Academia Programme.

{
    \small
    \bibliographystyle{ieeenat_fullname}
    \bibliography{main}

\begin{thebibliography}{38}
\providecommand{\natexlab}[1]{#1}
\providecommand{\url}[1]{\texttt{#1}}
\expandafter\ifx\csname urlstyle\endcsname\relax
  \providecommand{\doi}[1]{doi: #1}\else
  \providecommand{\doi}{doi: \begingroup \urlstyle{rm}\Url}\fi

\bibitem[Adam~Coates(2011)]{stl10}
Andrew Y.~Ng Adam~Coates, Honglak~Lee.
\newblock An analysis of single layer networks in unsupervised feature learning.
\newblock \emph{AISTATS}, 2011.

\bibitem[Birhane et~al.(2021)Birhane, Prabhu, and Kahembwe]{unsafe-data}
Abeba Birhane, Vinay~Uday Prabhu, and Emmanuel Kahembwe.
\newblock Multimodal datasets: misogyny, pornography, and malignant stereotypes.
\newblock \emph{ArXiv}, abs/2110.01963, 2021.

\bibitem[Bossard et~al.(2014)Bossard, Guillaumin, and Van~Gool]{food101}
Lukas Bossard, Matthieu Guillaumin, and Luc Van~Gool.
\newblock Food-101 -- mining discriminative components with random forests.
\newblock In \emph{European Conference on Computer Vision}, 2014.

\bibitem[Cao and Yang(2015)]{mu-origins}
Yinzhi Cao and Junfeng Yang.
\newblock Towards making systems forget with machine unlearning.
\newblock In \emph{2015 IEEE Symposium on Security and Privacy}, pages 463--480, 2015.

\bibitem[Chen et~al.(2021)Chen, Xie, and He]{vits}
Xinlei Chen, Saining Xie, and Kaiming He.
\newblock An empirical study of training self-supervised vision transformers.
\newblock In \emph{Proceedings of the IEEE/CVF International Conference on Computer Vision (ICCV)}, pages 9640--9649, 2021.

\bibitem[Chopra et~al.(2005)Chopra, Hadsell, and LeCun]{contrastive-loss}
S. Chopra, R. Hadsell, and Y. LeCun.
\newblock Learning a similarity metric discriminatively, with application to face verification.
\newblock In \emph{2005 IEEE Computer Society Conference on Computer Vision and Pattern Recognition (CVPR'05)}, pages 539--546 vol. 1, 2005.

\bibitem[Cooper et~al.(2024)Cooper, Choquette{-}Choo, Bogen, Jagielski, Filippova, Liu, Chouldechova, Hayes, Huang, Mireshghallah, Shumailov, Triantafillou, Kairouz, Mitchell, Liang, Ho, Choi, Koyejo, Delgado, Grimmelmann, Shmatikov, Sa, Barocas, Cyphert, Lemley, danah boyd, Vaughan, Brundage, Bau, Neel, Jacobs, Terzis, Wallach, Papernot, and Lee]{mu-doesnt-do}
A.~Feder Cooper, Christopher~A. Choquette{-}Choo, Miranda Bogen, Matthew Jagielski, Katja Filippova, Ken~Ziyu Liu, Alexandra Chouldechova, Jamie Hayes, Yangsibo Huang, Niloofar Mireshghallah, Ilia Shumailov, Eleni Triantafillou, Peter Kairouz, Nicole Mitchell, Percy Liang, Daniel~E. Ho, Yejin Choi, Sanmi Koyejo, Fernando Delgado, James Grimmelmann, Vitaly Shmatikov, Christopher~De Sa, Solon Barocas, Amy Cyphert, Mark Lemley, danah boyd, Jennifer~Wortman Vaughan, Miles Brundage, David Bau, Seth Neel, Abigail~Z. Jacobs, Andreas Terzis, Hanna~M. Wallach, Nicolas Papernot, and Katherine Lee.
\newblock Machine unlearning doesn't do what you think: Lessons for generative {AI} policy, research, and practice.
\newblock \emph{CoRR}, abs/2412.06966, 2024.

\bibitem[Desai et~al.(2021)Desai, Kaul, Aysola, and Johnson]{redcaps}
Karan Desai, Gaurav Kaul, Zubin Aysola, and Justin Johnson.
\newblock {RedCaps: Web-curated image-text data created by the people, for the people}.
\newblock In \emph{Advances in Neural Information Processing Systems Datasets and Benchmarks Track (Round 1)}, 2021.

\bibitem[Desai et~al.(2023)Desai, Nickel, Rajpurohit, Johnson, and Vedantam]{meru}
Karan Desai, Maximilian Nickel, Tanmay Rajpurohit, Justin Johnson, and Ramakrishna Vedantam.
\newblock Hyperbolic image-text representations.
\newblock In \emph{Proceedings of the 40th International Conference on Machine Learning}. JMLR.org, 2023.

\bibitem[{European Parliament} and {Council of the European Union}()]{gdpr}
{European Parliament} and {Council of the European Union}.
\newblock Regulation ({EU}) 2016/679 of the {European} {Parliament} and of the {Council}.

\bibitem[Gandikota et~al.(2023)Gandikota, Materzyńska, Fiotto-Kaufman, and Bau]{concept_prune}
Rohit Gandikota, Joanna Materzyńska, Jaden Fiotto-Kaufman, and David Bau.
\newblock Erasing concepts from diffusion models.
\newblock In \emph{2023 IEEE/CVF International Conference on Computer Vision (ICCV)}, pages 2426--2436, 2023.

\bibitem[Ganea et~al.(2018)Ganea, Becigneul, and Hofmann]{cones}
Octavian Ganea, Gary Becigneul, and Thomas Hofmann.
\newblock Hyperbolic entailment cones for learning hierarchical embeddings.
\newblock In \emph{Proceedings of the 35th International Conference on Machine Learning}, pages 1646--1655. PMLR, 2018.

\bibitem[Guo et~al.(2022{\natexlab{a}})Guo, Guo, and Yu]{cosne}
Yunhui Guo, Haoran Guo, and Stella Yu.
\newblock Co-sne: Dimensionality reduction and visualization for hyperbolic data.
\newblock pages 11--20, 2022{\natexlab{a}}.

\bibitem[Guo et~al.(2022{\natexlab{b}})Guo, Wang, Chen, and Yu]{hyperclass}
Yunhui Guo, Xudong Wang, Yubei Chen, and Stella~X. Yu.
\newblock Clipped hyperbolic classifiers are super-hyperbolic classifiers.
\newblock In \emph{2022 IEEE/CVF Conference on Computer Vision and Pattern Recognition (CVPR)}, pages 1--10, 2022{\natexlab{b}}.

\bibitem[Kingma and Ba(2014)]{adam}
Diederik Kingma and Jimmy Ba.
\newblock Adam: A method for stochastic optimization.
\newblock \emph{International Conference on Learning Representations}, 2014.

\bibitem[Kravets and Namboodiri(2025{\natexlab{a}})]{zeroshotunlearnlip}
Alexey Kravets and Vinay Namboodiri.
\newblock Zero-shot class unlearning in clip with synthetic samples.
\newblock In \emph{Proceedings of the Winter Conference on Applications of Computer Vision (WACV)}, pages 6456--6464, 2025{\natexlab{a}}.

\bibitem[Kravets and Namboodiri(2025{\natexlab{b}})]{zeroshotunlearn-spaceadapt}
Alexey Kravets and Vinay~P. Namboodiri.
\newblock Zero-shot {CLIP} class forgetting via text-image space adaptation.
\newblock \emph{Transactions on Machine Learning Research}, 2025{\natexlab{b}}.

\bibitem[Krizhevsky et~al.({\natexlab{a}})Krizhevsky, Nair, and Hinton]{cifar10}
Alex Krizhevsky, Vinod Nair, and Geoffrey Hinton.
\newblock Cifar-10 (canadian institute for advanced research).
\newblock {\natexlab{a}}.

\bibitem[Krizhevsky et~al.({\natexlab{b}})Krizhevsky, Nair, and Hinton]{cifar100}
Alex Krizhevsky, Vinod Nair, and Geoffrey Hinton.
\newblock Cifar-100 (canadian institute for advanced research).
\newblock {\natexlab{b}}.

\bibitem[Kumari et~al.(2023)Kumari, Zhang, Wang, Shechtman, Zhang, and Zhu]{ablating_concepts}
Nupur Kumari, Bingliang Zhang, Sheng-Yu Wang, Eli Shechtman, Richard Zhang, and Jun-Yan Zhu.
\newblock Ablating concepts in text-to-image diffusion models.
\newblock In \emph{2023 IEEE/CVF International Conference on Computer Vision (ICCV)}, pages 22634--22645, 2023.

\bibitem[Li et~al.(2024)Li, Chen, Xu, and Hu]{ood}
Huimin Li, Zhentao Chen, Yunhao Xu, and Junlin Hu.
\newblock Hyperbolic anomaly detection.
\newblock In \emph{Proceedings of the IEEE/CVF Conference on Computer Vision and Pattern Recognition (CVPR)}, pages 17511--17520, 2024.

\bibitem[Lin et~al.(2014)Lin, Maire, Belongie, Hays, Perona, Ramanan, Doll{\'a}r, and Zitnick]{coco}
Tsung-Yi Lin, Michael Maire, Serge Belongie, James Hays, Pietro Perona, Deva Ramanan, Piotr Doll{\'a}r, and C.~Lawrence Zitnick.
\newblock Microsoft coco: Common objects in context.
\newblock In \emph{Computer Vision -- ECCV 2014}, pages 740--755, Cham, 2014. Springer International Publishing.

\bibitem[Mettes et~al.(2024)Mettes, Ghadimi~Atigh, Keller-Ressel, Gu, and Yeung]{hyp-sur-cv}
Pascal Mettes, Mina Ghadimi~Atigh, Martin Keller-Ressel, Jeffrey Gu, and Serena Yeung.
\newblock Hyperbolic deep learning in computer vision: A survey.
\newblock \emph{International Journal of Computer Vision}, 132\penalty0 (9):\penalty0 3484--3508, 2024.

\bibitem[Nickel and Kiela(2017)]{poincare-emb}
Maximillian Nickel and Douwe Kiela.
\newblock Poincar\'{e} embeddings for learning hierarchical representations.
\newblock In \emph{Advances in Neural Information Processing Systems}. Curran Associates, Inc., 2017.

\bibitem[Nilsback and Zisserman(2008)]{flowers102}
Maria-Elena Nilsback and Andrew Zisserman.
\newblock Automated flower classification over a large number of classes.
\newblock In \emph{Indian Conference on Computer Vision, Graphics and Image Processing}, 2008.

\bibitem[Parkhi et~al.(2012)Parkhi, Vedaldi, Zisserman, and Jawahar]{pets}
Omkar~M. Parkhi, Andrea Vedaldi, Andrew Zisserman, and C.~V. Jawahar.
\newblock Cats and dogs.
\newblock In \emph{IEEE Conference on Computer Vision and Pattern Recognition}, 2012.

\bibitem[Peng et~al.(2022)Peng, Varanka, Mostafa, Shi, and Zhao]{hyp-sur}
Wei Peng, Tuomas Varanka, Abdelrahman Mostafa, Henglin Shi, and Guoying Zhao.
\newblock Hyperbolic deep neural networks: A survey.
\newblock \emph{IEEE Transactions on Pattern Analysis and Machine Intelligence}, 44\penalty0 (12):\penalty0 10023--10044, 2022.

\bibitem[Poppi et~al.(2025)Poppi, Poppi, Cocchi, Cornia, Baraldi, and Cucchiara]{safe-clip}
Samuele Poppi, Tobia Poppi, Federico Cocchi, Marcella Cornia, Lorenzo Baraldi, and Rita Cucchiara.
\newblock Safe-clip: Removing nsfw concepts from vision-and-language models.
\newblock In \emph{Computer Vision -- ECCV 2024}, pages 340--356, Cham, 2025. Springer Nature Switzerland.

\bibitem[Qian and Hu(2024)]{clip-zero-shot}
Qi Qian and Juhua Hu.
\newblock Online zero-shot classification with clip.
\newblock In \emph{Computer Vision -- ECCV 2024}, pages 462--477, Cham, 2024. Springer Nature Switzerland.

\bibitem[Radford et~al.(2021)Radford, Kim, Hallacy, Ramesh, Goh, Agarwal, Sastry, Askell, Mishkin, Clark, Krueger, and Sutskever]{clip}
Alec Radford, Jong~Wook Kim, Chris Hallacy, Aditya Ramesh, Gabriel Goh, Sandhini Agarwal, Girish Sastry, Amanda Askell, Pamela Mishkin, Jack Clark, Gretchen Krueger, and Ilya Sutskever.
\newblock Learning transferable visual models from natural language supervision.
\newblock In \emph{Proceedings of the 38th International Conference on Machine Learning}, pages 8748--8763. PMLR, 2021.

\bibitem[Radosavovic et~al.(2023)Radosavovic, Xiao, James, Abbeel, Malik, and Darrell]{clip-robot}
Ilija Radosavovic, Tete Xiao, Stephen James, Pieter Abbeel, Jitendra Malik, and Trevor Darrell.
\newblock Real-world robot learning with masked visual pre-training.
\newblock In \emph{Proceedings of The 6th Conference on Robot Learning}, pages 416--426. PMLR, 2023.

\bibitem[Ramesh et~al.(2022)Ramesh, Dhariwal, Nichol, Chu, and Chen]{clip-image}
Aditya Ramesh, Prafulla Dhariwal, Alex Nichol, Casey Chu, and Mark Chen.
\newblock Hierarchical text-conditional image generation with clip latents.
\newblock \emph{ArXiv}, abs/2204.06125, 2022.

\bibitem[Sala et~al.(2018)Sala, De~Sa, Gu, and Re]{hyp-trade}
Frederic Sala, Chris De~Sa, Albert Gu, and Christopher Re.
\newblock Representation tradeoffs for hyperbolic embeddings.
\newblock In \emph{Proceedings of the 35th International Conference on Machine Learning}, pages 4460--4469. PMLR, 2018.

\bibitem[van~den Oord et~al.(2018)van~den Oord, Li, and Vinyals]{infonce}
A{\"a}ron van~den Oord, Yazhe Li, and Oriol Vinyals.
\newblock Representation learning with contrastive predictive coding.
\newblock \emph{ArXiv}, abs/1807.03748, 2018.

\bibitem[van~der Maaten and Hinton(2008)]{tsne}
Laurens van~der Maaten and Geoffrey Hinton.
\newblock Visualizing data using t-sne.
\newblock \emph{Journal of Machine Learning Research}, 9\penalty0 (86):\penalty0 2579--2605, 2008.

\bibitem[Wang et~al.(2025)Wang, Lu, Zhang, Boenisch, Dziedzic, Yu, and Gao]{ac}
Yihan Wang, Yiwei Lu, Guojun Zhang, Franziska Boenisch, Adam Dziedzic, Yaoliang Yu, and Xiao-Shan Gao.
\newblock Machine unlearning for contrastive learning under auditing, 2025.

\bibitem[Xu et~al.(2023)Xu, Zhu, Zhang, Zhou, and Yu]{mu-survey}
Heng Xu, Tianqing Zhu, Lefeng Zhang, Wanlei Zhou, and Philip~S. Yu.
\newblock Machine unlearning: A survey.
\newblock \emph{ACM Comput. Surv.}, 56\penalty0 (1), 2023.

\bibitem[Yang et~al.(2024)Yang, Dai, Liu, Wang, Jiang, Tian, and Zhang]{cliperase}
Tianyu Yang, Lisen Dai, Zheyuan Liu, Xiangqi Wang, Meng Jiang, Yapeng Tian, and Xiangliang Zhang.
\newblock Cliperase: Efficient unlearning of visual-textual associations in {CLIP}.
\newblock \emph{CoRR}, abs/2410.23330, 2024.

\end{thebibliography}
}

\clearpage
\setcounter{page}{1}
\maketitlesupplementary

This supplementary material provides additional technical details, extended analyses, and supporting evidence for our main paper on machine unlearning in hyperbolic versus Euclidean contrastive learning spaces. We first present formal mathematical descriptions of the CLIP and MERU objectives to establish the geometric foundations that differentiate these approaches (\Cref{model-ojectives}). Next, we clarify the composition of our forget sets, highlighting the challenges in precisely defining concept boundaries (\Cref{forget-set}). We then provide comprehensive visualizations through confusion matrices that illustrate the different unlearning behaviors between models (\Cref{confusion-sup}). Additionally, we present complementary linear probing results that further confirm how feature representations are still linearly separable, which already can be observed in latent visualizations (\Cref{linear-probing}). Finally, we include extended visualizations of the latent spaces using multiple dimensionality reduction techniques to further support our findings on hyperbolic unlearning (\Cref{extra-sup}). These materials provide deeper technical understanding and additional empirical support for the conclusions presented in our main paper.

\section{Model Objectives}\label{model-ojectives}

\subsection{CLIP: Contrastive Learning in Euclidean Space}\label{clip-sup}
CLIP \cite{clip} consists of two encoders, a visual encoder $f_{\text{img}}$ and text encoder $f_{\text{txt}}$, mapping images and text into a shared Euclidean space $\mathbb{R}^d$. Given a batch of images and texts $\{(x_i, t_i)\}_{i=1}^N$, we obtain embeddings $x_i':=f_{\text{img}}(x_i)$ and $ t_i':=(f_{\text{txt}} t_i)$. CLIP is trained extending \ref{eq:infonce} to a symmetric cross-entropy loss:
\begin{align}
    L_{\text{CLIP}} =& -\frac{1}{2N}\sum_{i=1}^{N} \Bigg[ \log \frac{\exp(\text{sim}(x'_i, t'_i)/\tau)}{\sum_{j=1}^{N} \exp(\text{sim}(x'_i, t'_j)/\tau)} \\ 
    & + \log \frac{\exp(\text{sim}(x'_i, t'_i)/\tau)}{\sum_{j=1}^{N} \exp(\text{sim}(x'_j, t'_i)/\tau)}\Bigg].
\end{align}

where $\text{sim}(x'_i, t'_j) := \cos(\theta_{ij})$ is the cosine similarity between normalized image and text embeddings, $\theta_{ij}$ is the angle between them, and $\tau$ is a temperature parameter. This contrastive objective places all embeddings on a unit hypersphere, treating all concept relationships uniformly, regardless of their hierarchical nature.

\subsection{MERU: Contrastive Learning in Hyperbolic Space}\label{meru-pre}

MERU \cite{meru} extends contrastive learning to hyperbolic space using Lorentz model. MERU consists of visual and textual encoders, but it projects the image and text embeddings onto a hyperboloid manifold.
The distance between two points $x, y$ in the hyperboloid is given by 

\begin{equation}
    d_{\mathcal{L}}(x, y) = \frac{1}{\sqrt{c}} \cosh^{-1}(-c \langle x, y \rangle_{\mathcal{L}}),
\end{equation}

where 

\begin{equation}
    \langle x, y \rangle_{\mathcal{L}} = \langle x_{\text{space}}, y_{\text{space}}\rangle - x_{\text{time}} \cdot y_{\text{time}}
\end{equation}

is the Lorentzian inner product, $x = ( x_{\text{space}}, x_{\text{time}}) \in \mathbb{R}^{n+1}$, $x_{\text{space}} \in \mathbb{R}^{n}$ and $x_{\text{time}}\in\mathbb{R}$, and $c>0$ is the curvature of the space. The Lorentzian norm is defined by $||x||_{\mathcal{L}} = \sqrt{|\langle x, x \rangle_{\mathcal{L}}|}$. With this, the Lorentz model of curvature $-c$, $c>0$, and dimension $n$ is given by the set of vectors:

\begin{equation}
    \mathcal{L}^n:=\{x\in\mathbb{R}^{n+1}\;:\; \langle x, x \rangle_\mathcal{L} = -1/c\}.
\end{equation}

MERU is trained with a contrastive loss similar to CLIP, but using negative hyperbolic distance as the similarity measure, $\text{sim}_{\mathcal{L}}(x, y) = -d_{\mathcal{L}}(x, y)$.
Additionally, MERU incorporates an entailment loss to enforce partial order relationships between text and image embeddings:
\begin{equation}\label{meru:entail}
    L_{\text{entail}}(x, t) = \max(0, \text{ext}(x, t) - \text{aper}(t))
\end{equation}
where $\text{ext}(x, t)$ is the exterior angle between the text embedding $t$, given by
\begin{equation}
    \text{ext}(x,t) = \cos^{-1}\bigg(\frac{x_\text{time} + t_\text{time} \; c \; \langle x, t \rangle_\mathcal{L}}{\|t_\text{space}\|\sqrt{(c \; \langle x, t \rangle_\mathcal{L})^2 - 1}}\bigg),
\end{equation}
and image embedding $x$, and $\text{aper}(t)$ is the half-aperture of the entailment cone for $t$, 
\begin{equation}
    \text{aper}(t) = \sin^{-1}\Big(\frac{2K}{\sqrt{c}\|t_\text{space}\|}\Big).
\end{equation}

The hyperbolic geometry of MERU naturally accommodates hierarchical relationships, as the volume of the space grows exponentially with distance from the origin. This property allows generic concepts to be placed closer to the origin with more capacity to connect to numerous specific instances, in contrast to the uniform treatment of relationships in Euclidean space.

\section{Defining the forget set}\label{forget-set}

We built the different forget set aggregating related subfolders (e.g., \textit{dog = \{bordercollie, bostonterrier, etc.\}}, see Table \ref{tab:subreddits}). However, other subfolders in the retain set, such as \textit{alltheanimals}, may contain image-text pairs related to dogs, creating conflicting signals during unlearning. This highlights a broader challenge in the machine unlearning field: properly defining concept boundaries for removal remains an open problem \citep{mu-doesnt-do}.

\begin{table*}[h]
    \centering
    \renewcommand{\arraystretch}{1.2}
    \begin{tabular}{p{2cm}p{6cm}ccc}
    \toprule
         \textbf{Concept-class} & \textbf{Subreddits} & \textbf{Image-text samples} & \textbf{\% on Redcaps} & \textbf{\% on Redcaps2} \\
         \hline 
         dogs & \raggedright\small{dogpictures, bordercollie, bostonterrier, lookatmydog, doggos, bulldogs, australiancattledog, frenchbulldogs, bernesemountaindogs, australianshepherd, beagle, chihuahua, corgi, dobermanpinscher, husky, labrador, pitbulls, pomeranians, pug, pugs, rarepuppers, rottweiler} & 511585 & 4.26\% & 7.33\% \\
         \hline 
         cats & \raggedright\small{cats, blackcats, supermodelcats, catpictures, siamesecats, bengalcats, siberiancats} & 532640 & 4.43\% & 7.63\% \\
         \hline 
         food & \raggedright\small{food, foodporn, veganfoodporn, healthyfood, breakfastfood, chinesefood, tastyfood, budgetfood, baking, bento, breadit, breakfastfood, breakfast, burgers, chefit, pizza, sushi, tacos, veganrecipes, vegetarian} & 630971 & 5.25\% & 9.04\% \\
         \hline 
         plants & \raggedright\small{houseplants, plants, plantedtank, airplants, plantbaseddiet, plantsandpots, carnivorousplants, flowers, bonsai, botanicalporn, cactus, microgreens, monstera, orchids, permaculture, roses, succulents, vegetablegardening, gardening} & 587798 & 4.89\% & 8.42\% \\
         \hline 
         \textbf{Total} & 68 subreddits & 2262994 & 18.85\% & 32.43\% \\
    \bottomrule
    \end{tabular}
    \caption{Grouping of subreddits to higher-order concepts.}
    \label{tab:subreddits}
\end{table*}

\section{Confusion Matrices from Zero-Shot Classification}\label{confusion-sup}

\begin{figure*}[htbp]
   \centering
   \includegraphics[width=0.7\textwidth]{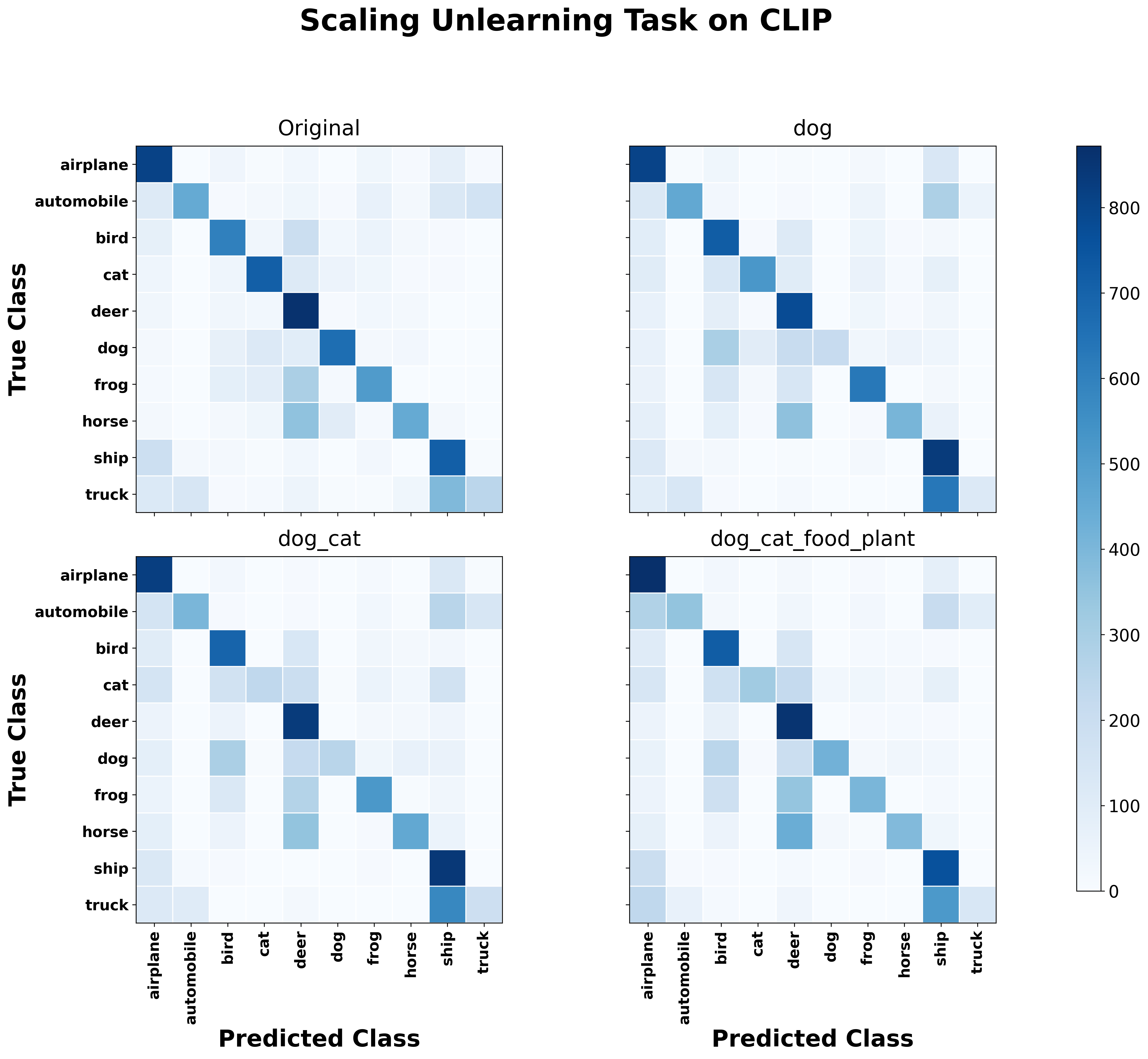}
   \caption{Confusion matrices for CLIP zero-shot classification at different scales of the unlearning task. After unlearning, CLIP shows moderate confusion, with dog images primarily misclassified as cats, but still retaining some dog classification capability.}
   \label{fig:clip_conf}
\end{figure*}

\begin{figure*}[htbp]
   \centering
   \includegraphics[width=0.7\textwidth]{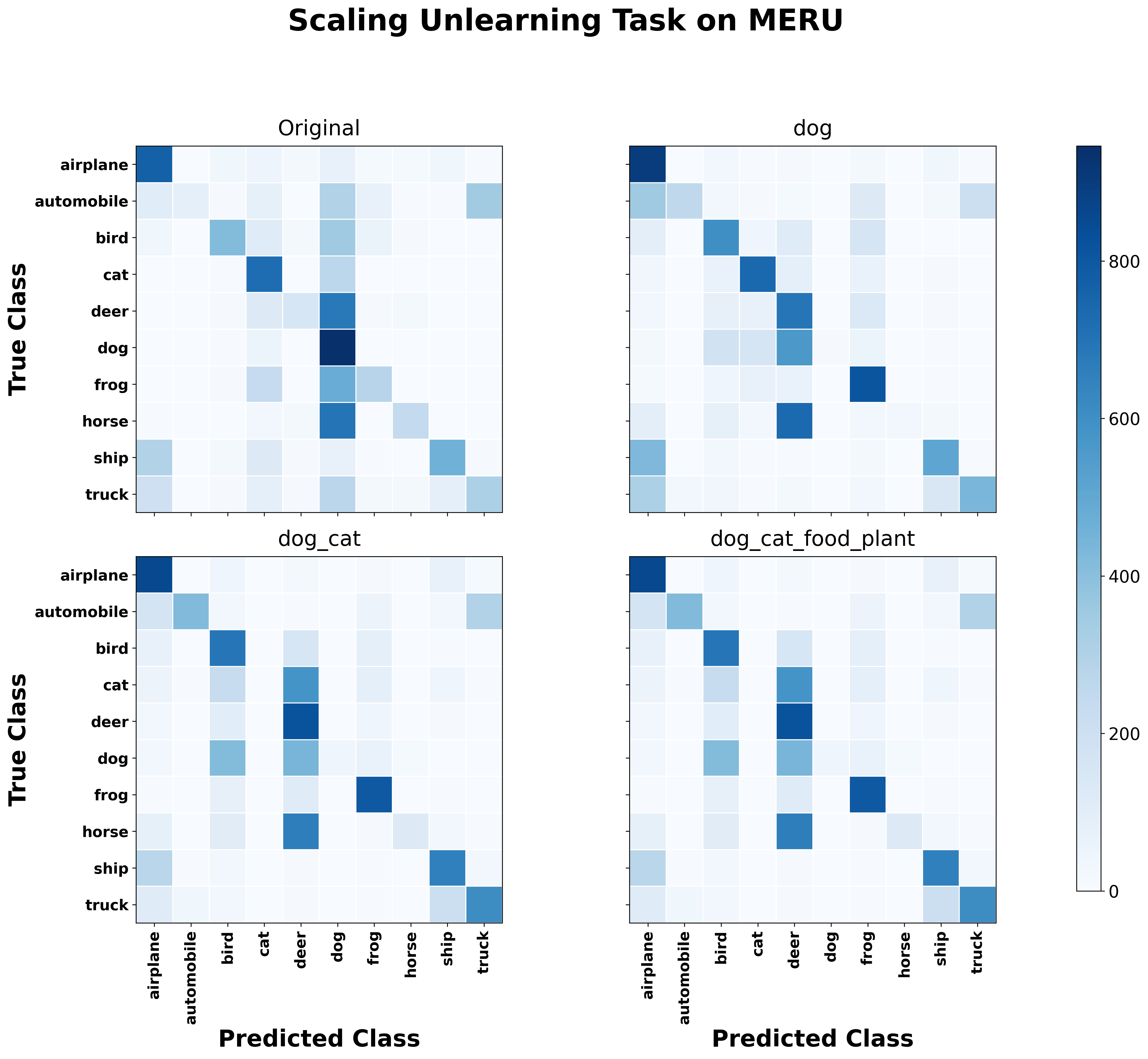}
   \caption{Confusion matrices for MERU zero-shot classification at different scales of the unlearning task. After unlearning, MERU demonstrates complete forgetting of the dog class, with dog images redistributed primarily to cat and horse categories according to semantic similarity.}
   \label{fig:meru_conf}
\end{figure*}

Figures \ref{fig:clip_conf} and \ref{fig:meru_conf} present confusion matrices for zero-shot classification before and after unlearning for CLIP and MERU, respectively. These visualizations provide detailed insights into how concept removal affects classification behavior across different classes.

For CLIP (Figure \ref{fig:clip_conf}), we observe partial concept removal, with the "dog" classification accuracy reduced but not eliminated. Most misclassified dog images are assigned to the "cat" category, indicating that CLIP maintains some understanding of semantic similarity even when attempting to forget. The retain classes show minimal disturbance, maintaining strong diagonal elements in the confusion matrix.

In contrast, MERU (Figure \ref{fig:meru_conf}) exhibits complete concept removal, with dog images almost entirely reassigned to other categories. The redistribution follows semantic hierarchies, with most dog images classified as "cat" which is a semantically animal category. This pattern supports our hypothesis that hyperbolic geometry leverages hierarchical relationships during unlearning, reassigning forgotten concepts according to their position in the semantic taxonomy. We further observe a decrease in performance on retaining the "horse" concept. However, this can be explained by observing how the original MERU already confuses horses by dogs, and then HAC, treating horses as if they were dogs, also removes them.

\section{Linear Probing}\label{linear-probing}

Linear probing extracts embeddings from image encoder and trains a linear classifier on these features. This evaluates whether class information remains linearly separable in the latent space after unlearning. Accuracy is reported for both retained and forgotten classes, quantifying how successfully target concepts have been removed while preserving desired knowledge. Testing linear separability of image features provides different insights. While R-acc and F-acc in zero-shot classification measure the unlearning performance in the alignment between images and texts embeddings. Here we measure whether after the damage is done the image features have been mixed between different categories or not.

\begin{table*}[t]
\centering
\caption{Linear probing accuracy in retain set (R-acc) and forget set (F-acc), across different tasks, after unlearning: (A) "dog"; (B) "dog" and "cat"; (C) "dog", "cat", "food" and "plant". We report results for both CLIP and MERU after alignment calibration using the optimal configuration from Section~\ref{sec:ablation_ac}. Values in \textbf{bold} indicate whether AC or HAC performed better at retaining or unlearning across A, B and C.}
\label{tab:scaling_analysis}
\resizebox{\textwidth}{!}{
\begin{tabular}{lcccccccccccccc}
\specialrule{1.5pt}{0pt}{0pt}
\multirow{2}{*}{Task} & \multirow{2}{*}{Method} & \multirow{2}{*}{\begin{tabular}[c]{@{}c@{}}Unlearn\\Set\end{tabular}} & \multicolumn{2}{c}{CIFAR-10} & \multicolumn{2}{c}{CIFAR-100} & \multicolumn{2}{c}{STL-10} & \multicolumn{2}{c}{O-IIIT Pets} & \multicolumn{2}{c}{Food101} & \multicolumn{2}{c}{Flowers102} \\
\cmidrule(lr){4-5} \cmidrule(lr){6-7} \cmidrule(lr){8-9} \cmidrule(lr){10-11} \cmidrule(lr){12-13} \cmidrule(lr){14-15}
& & & R-acc & F-acc & R-acc & F-acc & R-acc & F-acc & R-acc & F-acc & R-acc & F-acc & R-acc & F-acc \\
\specialrule{1.5pt}{0pt}{0pt}
 & & A & \textbf{89.9} & \textbf{85.2} & \textbf{71.5} &-& \textbf{95.1} & \textbf{92.4} & \textbf{86.1} & 87.5 & \textbf{84.5} & - & \textbf{95.4} & - \\
 & & B & \textbf{95.8} & \textbf{91.3} & \textbf{71.6} &-& \textbf{95.8} & \textbf{91.3} & - & 87.3 & \textbf{84.6} & - & \textbf{95.7} & - \\
 & \multirow{-3}{*}{AC} & C & \textbf{95.9} & \textbf{91.4} & \textbf{71.0} & 84.3 & \textbf{95.9} & \textbf{91.4} & - & 87.0& - & 84.3 & - & 95.4 \\
 \cmidrule(l){2-15}
 & & A & 89.3 & 85.5 & 69.8 &-& 94.9 & 93.6 & 84.8 & \textbf{86.7} & 83.8 & - & 93.8 & - \\
 & & B & 95.5 & 92.1 & 69.7 &-& 95.5 & 92.1 & - & \textbf{85.0} & 83.9 & - & 93.7 & - \\
 \multirow{-6}{*}{\begin{tabular}[c]{@{}l@{}}Linear Probe\\Classification\end{tabular}}& \multirow{-3}{*}{HAC-reg}& C & 95.4 & 92.4 & 68.6 & \textbf{83.1} & 95.4 & 92.4 & - & \textbf{85.6} & - & \textbf{83.0} & - & \textbf{92.6} \\
\specialrule{1.5pt}{0pt}{0pt}
\end{tabular}
}
\end{table*}

The linear probe classification results provide complementary insights into how unlearning affects the underlying feature representations. Consistent with prior work on hyperbolic classification \citep{hyperclass}, Euclidean representations show slightly better linear separability. However, the high forget accuracies in both methods reveal an important distinction between our approach and traditional class-unlearning: alignment calibration specifically targets cross-modal associations rather than altering the fundamental feature structure of either modality in isolation. This explains why images from the class related to the concept to forget remain linearly separable—their visual features are preserved while their association with corresponding text is disrupted. This insight can be also illustrated in a qualitative analysis of the latent spaces \cref{sec:qual}.



\section{Additional Latent Space Visualizations}\label{extra-sup}

\begin{figure*}[htbp]
    \centering
    \begin{subfigure}[b]{0.33\textwidth}
        \centering
        \includegraphics[width=\textwidth]{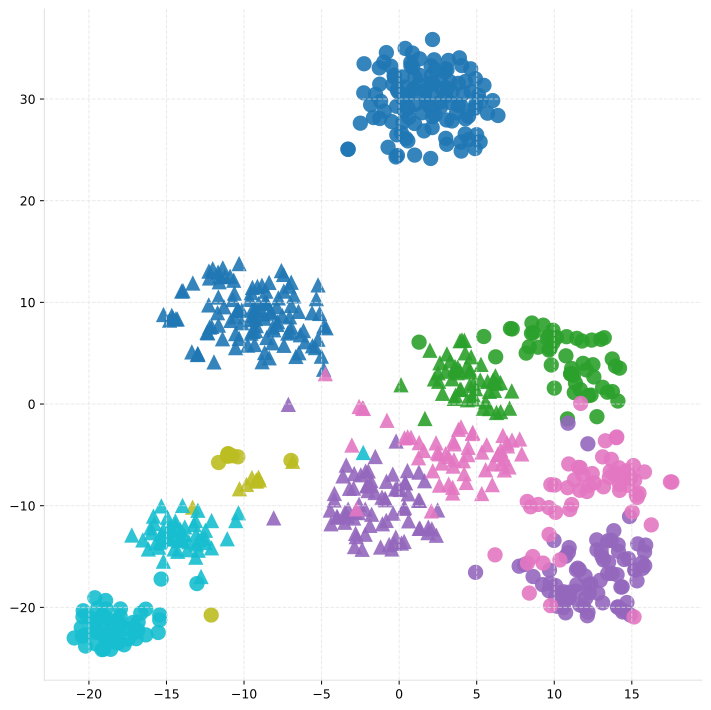}
        \caption{Original MERU: T-SNE}
        \label{fig:sub1}
    \end{subfigure}
    \hfill
    \begin{subfigure}[b]{0.33\textwidth}
        \centering
        \includegraphics[width=\textwidth]{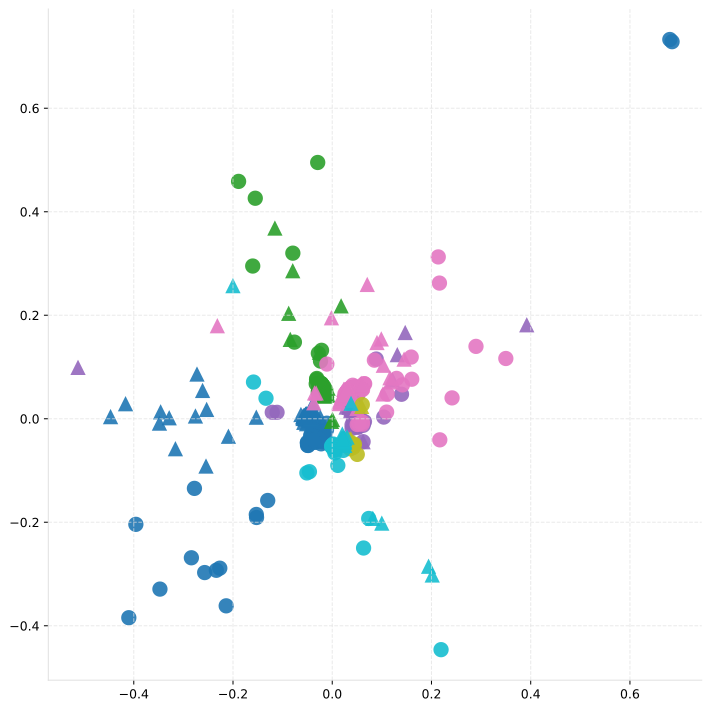}
        \caption{Original MERU: HT-SNE}
        \label{fig:sub2}
    \end{subfigure}
    \hfill
    \begin{subfigure}[b]{0.33\textwidth}
        \centering
        \includegraphics[width=\textwidth]{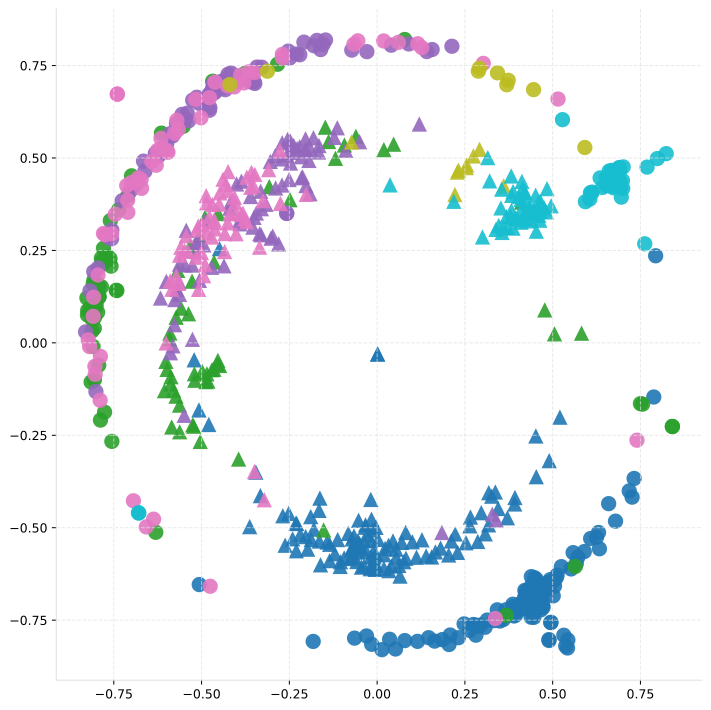}
        \caption{Original MERU: CO-SNE}
        \label{fig:sub3}
    \end{subfigure}
    \\
    \begin{subfigure}[b]{0.33\textwidth}
        \centering
        \includegraphics[width=\textwidth]{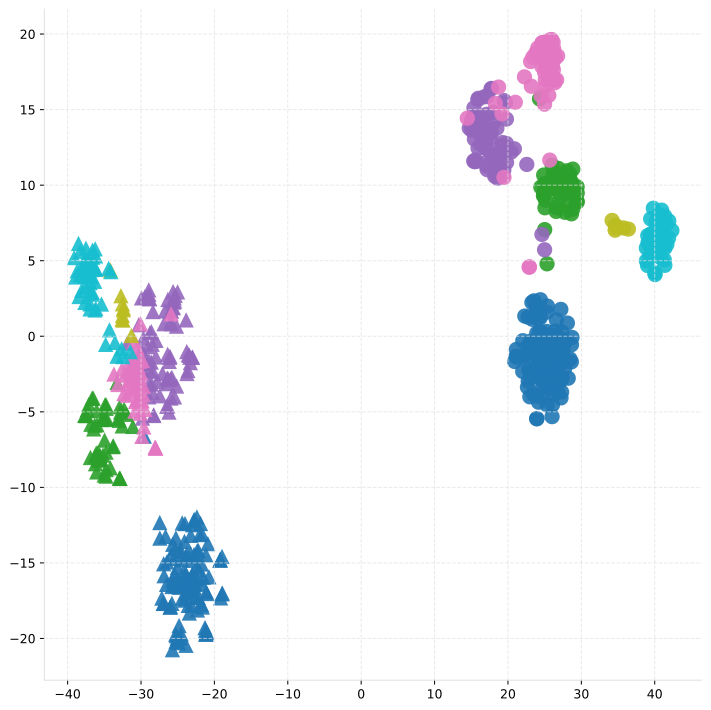}
        \caption{Unlearned MERU: T-SNE}
        \label{fig:sub1}
    \end{subfigure}
    \hfill
    \begin{subfigure}[b]{0.33\textwidth}
        \centering
        \includegraphics[width=\textwidth]{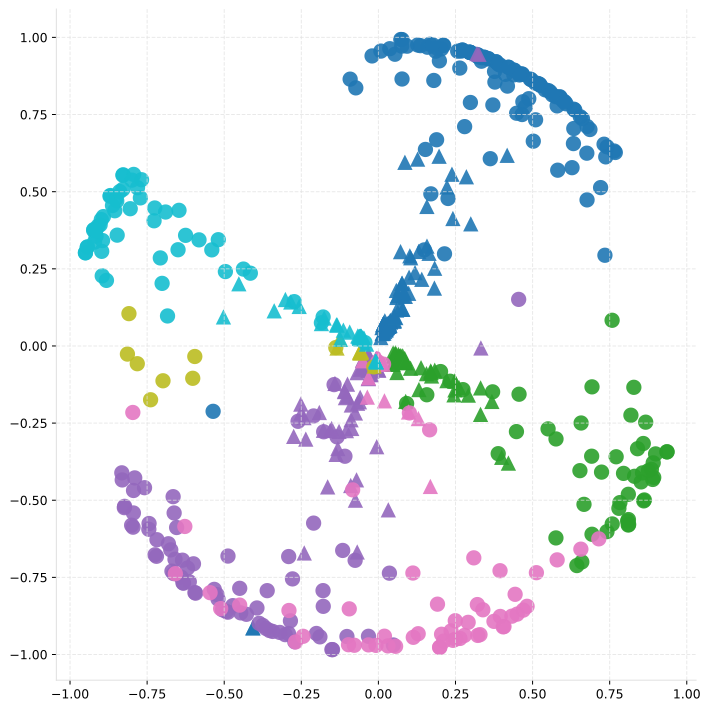}
        \caption{Unlearned MERU: HT-SNE}
        \label{fig:sub2}
    \end{subfigure}
    \hfill
    \begin{subfigure}[b]{0.33\textwidth}
        \centering
        \includegraphics[width=\textwidth]{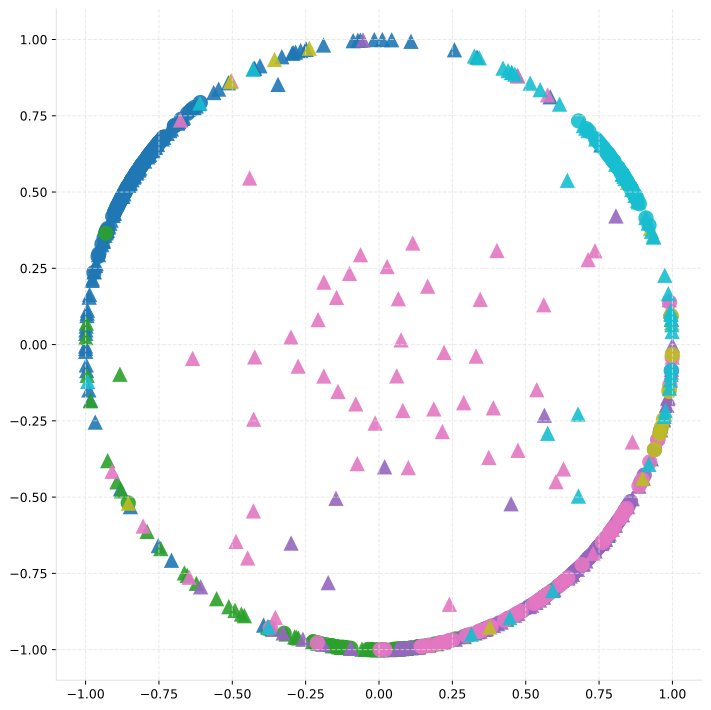}
        \caption{Unlearned MERU: CO-SNE}
        \label{fig:sub4}
    \end{subfigure}
    \caption{Latent space visualizations with T-SNE, hyperbolic T-SNE and CO-SNE of MERU before and after removing the concept-class "dog". $\triangle$ refer to text embeddings, $\circ$ to image embeddings, and colors to \textcolor{tabpink}{\textbf{dogs}}, \textcolor{tabpurple}{\textbf{cats}}, \textcolor{tabcyan}{\textbf{pizzas}}, \textcolor{tabblue}{\textbf{buses}}, \textcolor{tabgreen}{\textbf{birds}}, and \textcolor{tabolive}{\textbf{apples}}.}
    \label{extra-visual}
\end{figure*}

\begin{figure*}[htbp]
    \centering
    \begin{subfigure}[b]{0.45\textwidth}
        \centering
        \includegraphics[width=\textwidth]{figs/6_cosne_meru.png}
        \caption{Original MERU}
        \label{fig:sub1}
    \end{subfigure}
    \hfill
    \begin{subfigure}[b]{0.45\textwidth}
        \centering
        \includegraphics[width=\textwidth]{figs/6_cosne_u_meru.png}
        \caption{Unlearned "dog"}
        \label{fig:sub2}
    \end{subfigure}
    \\
    \begin{subfigure}[b]{0.45\textwidth}
        \centering
        \includegraphics[width=\textwidth]{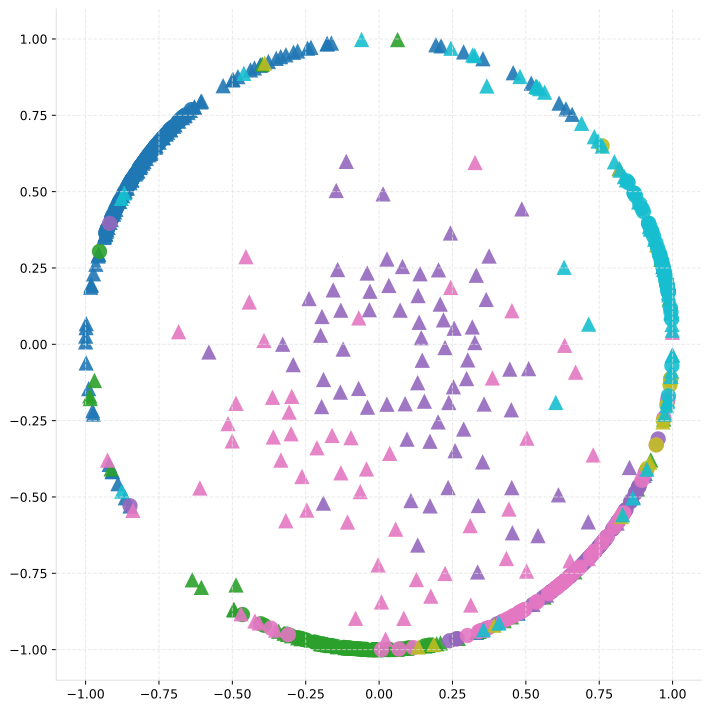}
        \caption{Unlearned "dog","cat"}
        \label{extra-visual-2:fig:sub1}
    \end{subfigure}
    \hfill
    \begin{subfigure}[b]{0.45\textwidth}
        \centering
        \includegraphics[width=\textwidth]{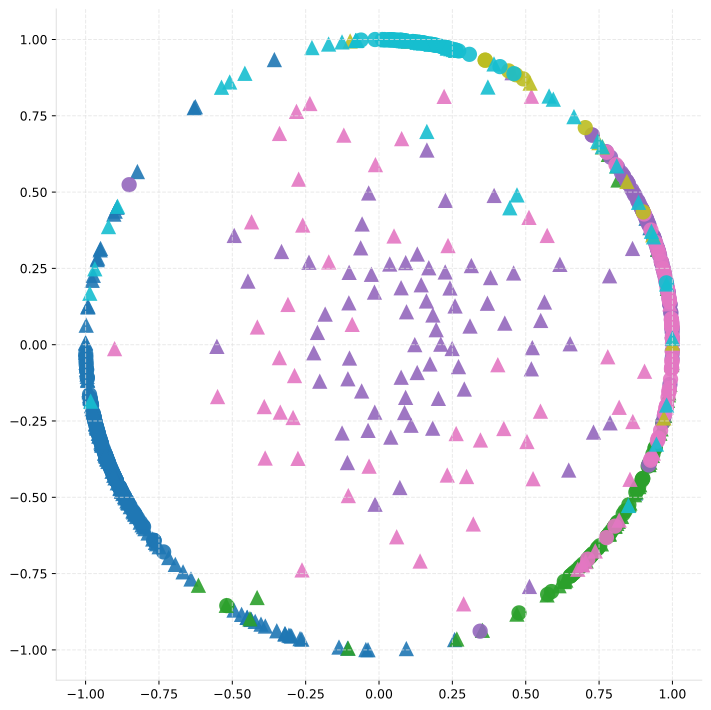}
        \caption{Unlearned "dog", "cat", "food", "plant"}
        \label{extra-visual-2:fig:sub2}
    \end{subfigure}
    \caption{Latent space visualizations with CO-SNE of MERU at different unlearning tasks. $\triangle$ refer to text embeddings, $\circ$ to image embeddings, and colors to \textcolor{tabpink}{\textbf{dogs}}, \textcolor{tabpurple}{\textbf{cats}}, \textcolor{tabcyan}{\textbf{pizzas}}, \textcolor{tabblue}{\textbf{buses}}, \textcolor{tabgreen}{\textbf{birds}}, and \textcolor{tabolive}{\textbf{apples}}.}
    \label{extra-visual-2}
\end{figure*}

To complement visualizations from \Cref{sec:qual} we include more instances in visualizations. Additionally, for the hyperbolic case, we include visualizations from another perspective, using CO-SNE \cite{cosne}. This method leverages hyperbolic Cauchy distribution (instead of hyperbolic student's t-distribution) and Lorentz distance, to represent global hierarchy and local distances in the same visualization. This allow us to "zoom-in" and see the origin of the hyperboloid from a closer point of view. \Cref{extra-visual}, illustrates the latent space of MERU from three perspectives before and after unlearning. CO-SNE allow us to better see that text embeddings of "dogs" remain closer to the origin, while other instances are pushed further, as discussed in \Cref{sec:qual}. \Cref{extra-visual-2} illustrates the same idea when scaling the unlearning problem to multiple concept removal. Observe that when "cats" are included in the forget set, the text embeddings for cats also remain close to the origin, and this is not disturb when including "food" and "plants", illustrating the robustness of HAC at scaling the unlearning task.

\end{document}